\documentclass[lettersize,journal]{IEEEtran}
\usepackage{amsmath,amsfonts}
\usepackage{algorithmic}
\usepackage{algorithm}
\usepackage{array}
\usepackage[caption=false,font=normalsize,labelfont=sf,textfont=sf]{subfig}
\usepackage{textcomp}
\usepackage{stfloats}
\usepackage{url}
\usepackage{verbatim}
\usepackage{graphicx}
\usepackage{cite}

\usepackage{amssymb}
\usepackage{booktabs}
\usepackage{multirow}
\usepackage{colortbl}
\usepackage{makecell}
\usepackage{lineno}
\usepackage{wrapfig}
\usepackage{enumitem}
\makeatletter
\renewcommand{\maketag@@@}[1]{\hbox{\m@th\normalsize\normalfont#1}}%
\makeatother

\begin{document}

\title{Transformer-based Multimodal Change Detection with Multitask Consistency Constraints}

\author{
	\IEEEauthorblockN{Biyuan Liu$^a$, Huaixin Chen$^{a}$, Kun Li$^b$, Michael Ying Yang$^{b*}$}\\
	\IEEEauthorblockA{$^a$ University of Electronic Science and Technology of China}\\
	\IEEEauthorblockA{$^b$ University of Twente}\\
	\IEEEauthorblockA{lby9469@gmail.com, huaixinchen@uestc.edu.cn, \{k.li, michael.yang\}@utwente.nl}
}

\markboth{}%
{Shell \MakeLowercase{\textit{et al.}}: A Sample Article Using IEEEtran.cls for IEEE Journals}


\maketitle

\begin{abstract}
Change detection plays a fundamental role in Earth observation for analyzing temporal iterations over time. However, recent studies have largely neglected the utilization of multimodal data that presents significant practical and technical advantages compared to single-modal approaches. This research focuses on leveraging {pre-event} digital surface model (DSM) data and {post-event} digital aerial images captured at different times for detecting change beyond 2D. We observe that the current change detection methods struggle with the multitask conflicts between semantic and height change detection tasks. To address this challenge, we propose an efficient Transformer-based network that learns shared representation between cross-dimensional inputs through cross-attention. {It adopts a consistency constraint to establish the multimodal relationship. Initially, pseudo-changes are derived by employing height change thresholding. Subsequently, the $L2$ distance between semantic and pseudo-changes within their overlapping regions is minimized. This explicitly endows the height change detection (regression task) and semantic change detection (classification task) with representation consistency.} A DSM-to-image multimodal dataset encompassing three cities in the Netherlands was constructed. It lays a new foundation for beyond-2D change detection from cross-dimensional inputs. Compared to five state-of-the-art change detection methods, our model demonstrates consistent multitask superiority in terms of semantic and height change detection. Furthermore, the consistency strategy can be seamlessly adapted to the other methods, yielding promising improvements.
\end{abstract}


\section{Introduction}
The field of change detection is undergoing a significant evolution characterized by higher temporal frequencies \cite{toker2022dynamicearthnet}, finer-grained analyses \cite{tian2022large}, and increased dimensionality \cite{cserep2023distributed,stilla2023change}. Recent advancements in Earth observation techniques have enabled daily change detection \cite{toker2022dynamicearthnet} and fine-grained analysis spanning up to nine distinct change categories \cite{tian2022large}. Moreover, there have been exciting breakthroughs beyond traditional 2D change detection \cite{cserep2023distributed,stilla2023change}. However, a predominant number of prevailing developments still center around single-modal and 2D change detection, such as introducing contrastive metrics for learning class-distinct features \cite{CD2017contra,zagoruyko2015learning}, leveraging multitask consistency for semi-supervised training \cite{quan2023unified,peng2020semicdnet}, and adopting attention mechanisms to model long-range context \cite{chen2021remote,zhang2022swinsunet}.

Some noteworthy examples demonstrate the incorporation of multimodal data into change detection task offers both practical flexibility and technical advantages. Combining optical images with synthetic aperture radar (SAR) \cite{touati2017energy,chen2022unsupervised}  alleviates weather-related and atmospheric restrictions. Using point cloud data from Lidar and photogrammetry \cite{zhang2019change} for detecting 3D changes is obviously less constrained in input pairs formation, resulting in enhanced flexible application. Furthermore, the different imaging modalities may be complementary for enhancing the change detection in some extreme conditions (e.g. {flooding} \cite{touati2019multimodal} and burned areas \cite{fodor2023rapid}). In our particular context, merging Digital Surface Models (DSM) with aerial imagery incorporates vital vertical data, enhancing the granularity of change detection. Our approach surpasses the conventional 2D semantic change detection methods by providing a more nuanced understanding of spatial changes (see Figure \ref{fig_dataset_sample}). 

\begin{figure*}[!th]
	\centering
	\includegraphics[width=1.0\textwidth]{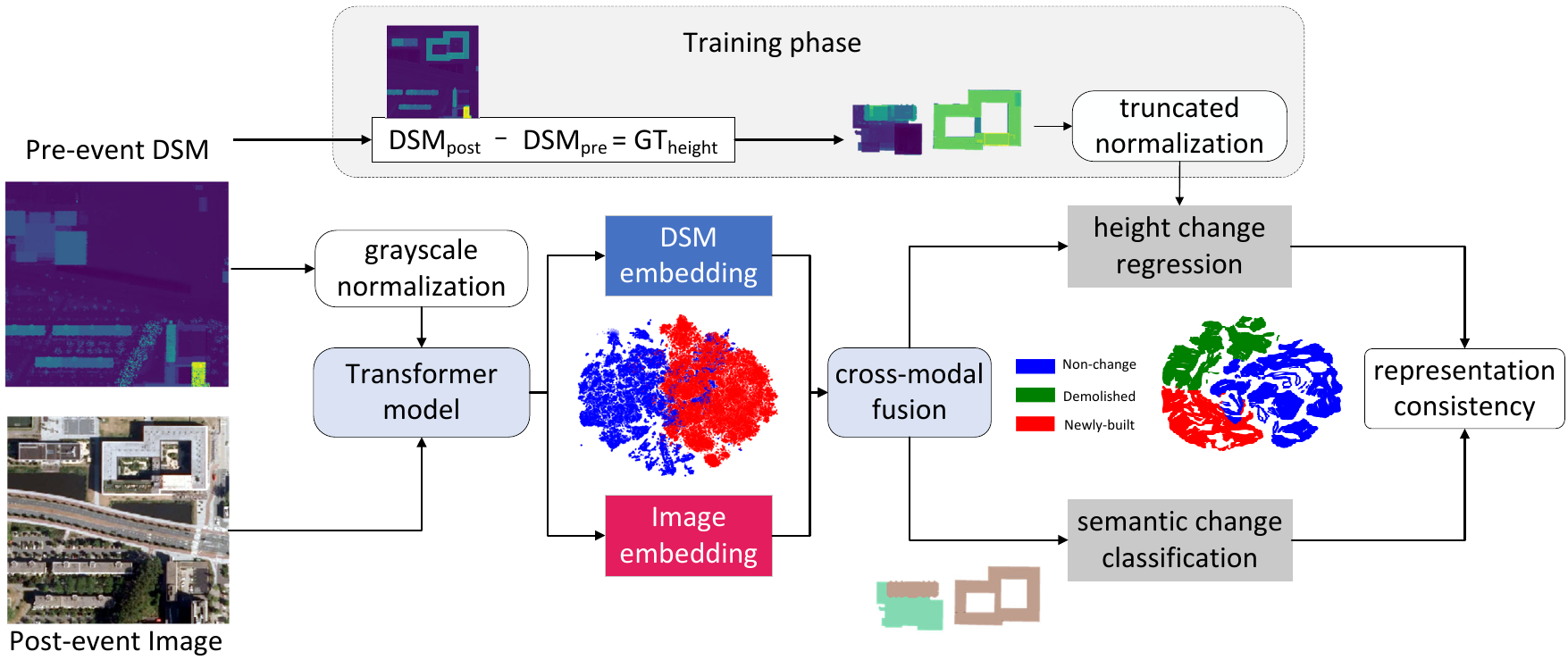}
	\caption{The conceptual pipeline showing how multimodal image and DSM data are utilized for detecting height and semantic changes simultaneously.}
	\label{fig:conceptual_pipeline}
\end{figure*}

Due to the scarcity of bi-temporal 3D data, existing methods for high dimensional change detection often {rely on} multi-source 3D data, {requiring} manual modality alignment before change detection, such as dense image matching \cite{zhang2022photogrammetric} and artificial feature selection \cite{zhang2019change, zhang2021novel}, which are time-consuming processes that risk information loss. In MTBIT \cite{marsocci2023inferring}, the use of bi-temporal 2D images to infer changes in building height overlooks the benefits of multimodal data integration, as exhaustively discussed in section 2. Consequently, there is an ongoing gap in research on change detection beyond 2D that fully leverages the potential of multimodal data.

\begin{figure}[tp]%
	\centering
	\includegraphics[width=0.5\textwidth]{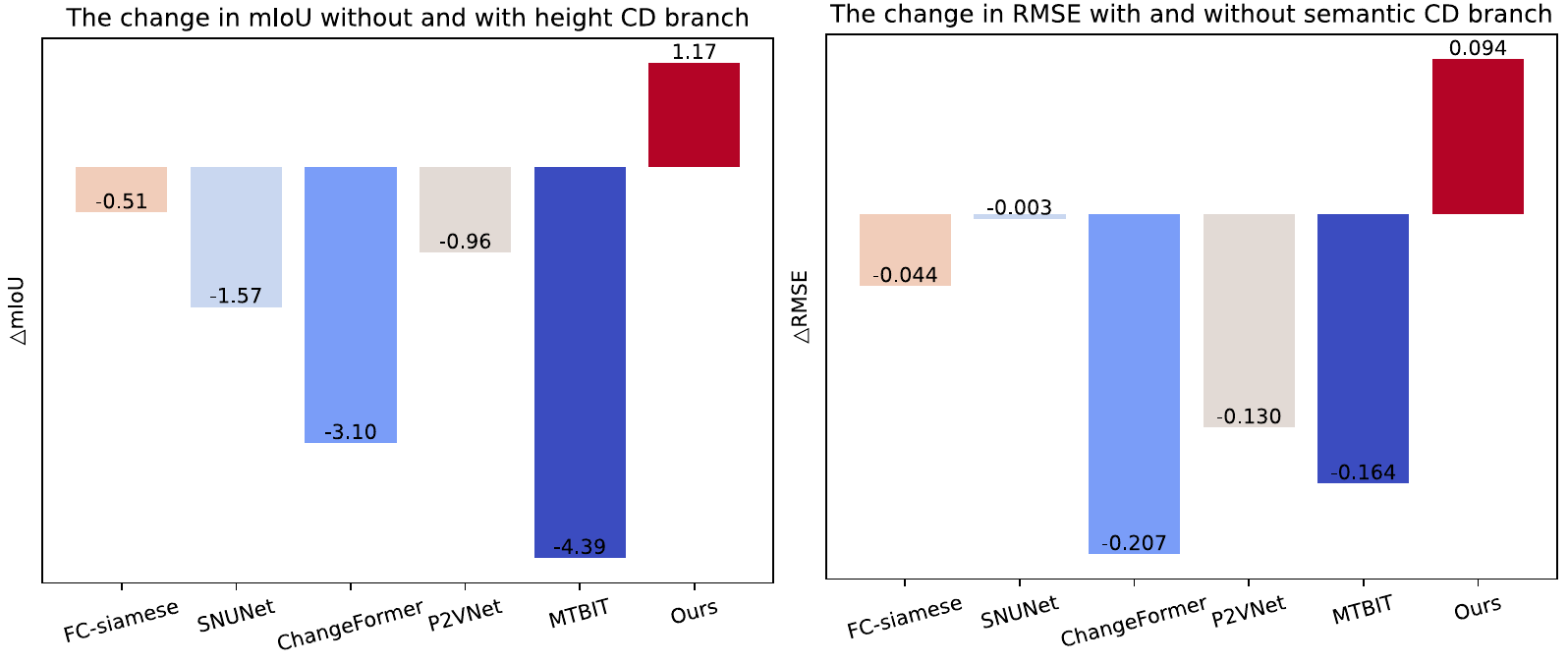}
	\caption{The performance change of semantic (left) and height (right) change detection in a single-task and multitask manner, which implies the multitask conflicts between 2D semantic and height change detection.}\label{fig1_compare}
\end{figure}

To address these gaps and tackle the inherent challenges, we have developed a multi-temporal dataset named Hi-BCD. This dataset comprises pairs of pre-event Digital Surface Models (DSM) representing height data and post-event aerial images. It is designed for detecting multi-category semantic and height changes in buildings across three cities in the Netherlands. Through extensive benchmarking of the state-of-the-art change detection methods, including convolutional neural network (CNN) based and Transformer-based \cite{vaswani2017attention} methods, we discover the potential multitask conflicts between semantic change detection (classification task) and height change detection (regression task). In Figure \ref{fig1_compare}, the multitask branches have brought great impact on each other. The performance of semantic change detection branch declined due to the height branch, while the height change detection branch conversely gained some improvements due to semantic hints. Therefore, we propose a novel Transformer-based pipeline that learns shared representation from images and DSM data via cross-attention (see Figure \ref{fig:conceptual_pipeline}). It is equipped with an explicit multitask consistency strategy, which involves the mapping from continuous height change to discrete pseudo change with a soft-thresholding module. Then the pixel-wise similarity is maximized between pseudo change and real semantic change, enabling the information interaction of multimodal change maps. The contributions of this paper are summarized as follows:
\begin{itemize}
	\item We propose an efficient and light Transformer-based network that {fuses feature of} cross-dimensional modalities via parallelly arranged cross-attention modules.
	\item We reveal the potential multitask conflicts in state-of-the-art methods while simultaneously handling semantic and height change detection. We propose a multitask consistency constraint that quantifies the similarity between semantic and pseudo change obtained through height change thresholding for alleviating multitask interference.
	\item We build a multimodal DSM-to-image {buiding change detection} dataset {called Hi-BCD,} with generously sized high-resolution tiles. It enables the detection of 2D semantic and 3D height changes simultaneously from cross-dimensional modalities. 
	\item The experiment in Hi-BCD demonstrates that our method outperforms existing methods with consistent semantic and height change detection results. Additionally, the proposed consistency strategy can be easily employed to enhance the other methods.
\end{itemize}

The rest of this paper is organized as follows: Section 2 provides a brief overview of single-modal and multimodal change detection methods. Section 3 introduces the proposed multimodal change detection network and multitask consistency. Section 4 describes our dataset. In Section 5, we perform a comparison with some current state-of-the-art convolutional neural network (CNN) based and Transformer-based change detection models, which reveals the multitask conflicts and demonstrates the superiority of our method. We conduct ablation studies about the influence of multitask consistency in semantic and height change detection, providing a better understanding of our model. Section 6 draws conclusions.

\section{Related Work}\label{sec2}

\textbf{Single-modal change detection.} The most remarkable achievements occur in the field of single-modal 2D image change detection, where large-scale data are available for utilization. These studies have improved the accuracy and efficiency with superior training metrics \cite{CD2017contra,zagoruyko2015learning}, densely-connected structure \cite{fang2021snunet,zhang2020deeply}, enhanced local and global context aggregation \cite{daudt2018fully,chen2021remote,quispe2020automatic}, light-weight components \cite{liu2022lsnet}, and decoupled change modeling \cite{lin2022transition}. Some latest studies also focus on detecting multi-class changes for in-depth scene understanding \cite{peng2021scdnet,tian2022large}. {In \cite{qi2018building}, Qi et al. introduced a grid-based method that categorizes grids into one of three change patterns: significant increase, significant decrease, or roughly unchanged, marking a substantial advancement beyond traditional binary outcome approaches (changed or unchanged).}

It is a significant trend that also challenges to detect the volumetric or vertical information in real applications such as quantitative estimation of changes in urban areas, forest biomasses, and land morphology \cite{qin20163d, marsocci2023inferring}. As multi-view imaging and aerial laser scanning (ALS) technologies continue to advance, an increasing amount of DSM-to-DSM \cite{cserep2023distributed} and cloud-to-cloud differencing methods \cite{qin20163d,de2023deep} have emerged. However, it is a strong hypothesis that multi-temporal data are available, especially for 3D data, leaving a barrier to the wide applicability of these methods.

\textbf{Multimodal change detection.}  A large number of multimodal change detection studies focus on detecting changes between optical and SAR images to alleviate the restriction of weather and atmosphere. Due to the scarcity of multimodal data, {the recent studies} tend to build the pixel-wise or graph correlation in an unsupervised manner without considering the deep features, such as the energy-based model \cite{touati2017energy}, coupled dictionary learning \cite{ferraris2019coupled},  Markov random field model \cite{touati2019multimodal}, change vector analysis \cite{chirakkal2022unsupervised}, and graph representation learning \cite{chen2022unsupervised, sun2022structured,sun2021iterative}. {Sun et al. \cite{sun2021iterative} presents an iterative robust graph combined with Markovian co-segmentation, focusing on structure consistency for enhanced detection accuracy. The \cite{sun2021structure,sun2022image} further this by employing structure cycle consistency and an improved nonlocal patch-based graph to address noise and sensor differences, showing superior performance across multiple datasets and scenarios. These methods signify advancements in unsupervised change detection without the need for labeled data.} Regarding these deep learning approaches, the majority employ an explicit image translation process \cite{luppino2021deep,li2021deep,wu2021commonality}. Conversely, some methods prefer a single-modal change detection framework that projects two heterogeneous images into a common latent space \cite{shao2021sunet,yang2021dpfl}.

Some recent efforts are delving into change detection beyond 2D with multimodal data. In \cite{zhang2018change}, the Siamese CNN was employed to detect changes between point clouds obtained from ALS and dense image matching. In \cite{zhang2021novel}, the features including color, shape, and elevation maps are manually extracted for change detection between the point cloud and image. In \cite{zhang2022photogrammetric}, a multi-source point cloud processing network was devised to detect genuine 3D changes. Yet, most of these methods require a time-consuming pre-processing step for modality alignment, which potentially results in information loss. Conversely, we directly handle multimodal data across different dimensions inspired by cross-attention mechanisms \cite{chen2021crossvit}.
In MTBIT \cite{marsocci2023inferring}, it attempted to infer a change map represented by DSM from bi-temporal 2D images. Unlike MTBIT, we propose to directly deal with DSM-to-image multimodal inputs for various reasons: 1) Estimating height from single view image remains an ill-posed problem. The introduced pre-temporal DSM provides abundant context priors about vertical information near the change areas. 2) The ground truth elevation change is essentially generated with bi-temporal DSMs, either in MTBIT or our method. Therefore, it is under-utilization of a considerable amount of 3D information in MTBIT. 3) For change detection that spans over a long time, there may exist a significant resolution gap in multi-temporal images (e.g., 2.0m vs. 0.25m as shown in Figure \ref{fig_dataset_sample}(a)(c)). On the contrary, the DSM data derived from point cloud in our dataset allows for at least 4 point records at a 0.25$\times$0.25$m^2$ grid. 

\textbf{Multitask learning.} 
In multi-task learning (MTL), a single model is trained to simultaneously predict outcomes for multiple tasks, leveraging data across these tasks to achieve better performance than if each task were learned independently \cite{liu2021conflict, sener2018multi, zhang2021survey}. Unfortunately, MTL often causes performance degradation compared to single-task models \cite{standley2020tasks}. A main reason for such degradation is gradients conflict \cite{wang2020gradient, liu2021conflict}. {The {\bf \textit{model-level}} multitask optimization involves addressing multitask conflicts through gradient manipulation. These per-task gradients may have conflicting directions or a large difference in magnitudes, with the largest gradient dominating the update direction.
	Various heuristics have been introduced for manipulating the task-specific gradient, such as the uncertainty of the tasks \cite{kendall2018multi}, the norm of the gradients \cite{chen2018gradnorm}, equal cosine similarities \cite{liu2021towards}, and Pareto optimal \cite{navon2022multi}.
	The {\bf \textit{Task-level}} multitask learning typically establishes correlations among multiple tasks using specific transformations \cite{zhu2020edge,kendall2018multi,zhang2019pattern} or by integrating multiple tasks that are inherently consistent \cite{zheng2022changemask,yang2021asymmetric,deng2022feature}. For instance, \cite{zhang2019pattern} learns a transformation between semantic segmentation and depth feature spaces. Zhu et al. \cite{zhu2020edge} explicitly measure the border consistency between segmentation and depth and minimize it in a greedy manner by iteratively supervising the network towards a locally optimal solution.  Kendall et al. \cite{kendall2018multi} models the uncertainty of segmentation
	and depth to re-weight themselves in the loss function. In the context of change detection, the auxilary task to predict the segmentation boundaries of bi-temporal inputs is widely adopted \cite{zheng2022changemask,yang2021asymmetric,deng2022feature}, where the learned boundary representation can be shared with the change detection branch. The auxiliary constraint is usually beneficial, as it introduces inductive bias through the inclusion of related additional information. Nonetheless, as reported in \cite{daudt2019multitask}, it can sometimes hamper the performance of each task. Certain studies have delved into the multitask relationship by considering multitask consistency. In \cite{tian2022large}, three consistency metrics including binary, change area, and no-change area consistency are used for evaluation. In \cite{shu2022mtcnet}, the consistency between bi-temporal semantic labels and the change labels is exploited to enhance semi-supervised generalization. 
	
	Zhu et al. \cite{zhu2020edge} underscore the unique challenge of correlating semantic maps with depth maps due to their significant but complex relationship. We explore a straightforward yet potent task-level transformation to navigate the intricacies of multitask conflicts. We highlight the connection between negatively changed height, often associated with demolished buildings, and positively changed height, as observed in newly constructed areas.  }

\section{Method}

\begin{figure*}[!th]%
	\centering
	\includegraphics[width=1.0\textwidth]{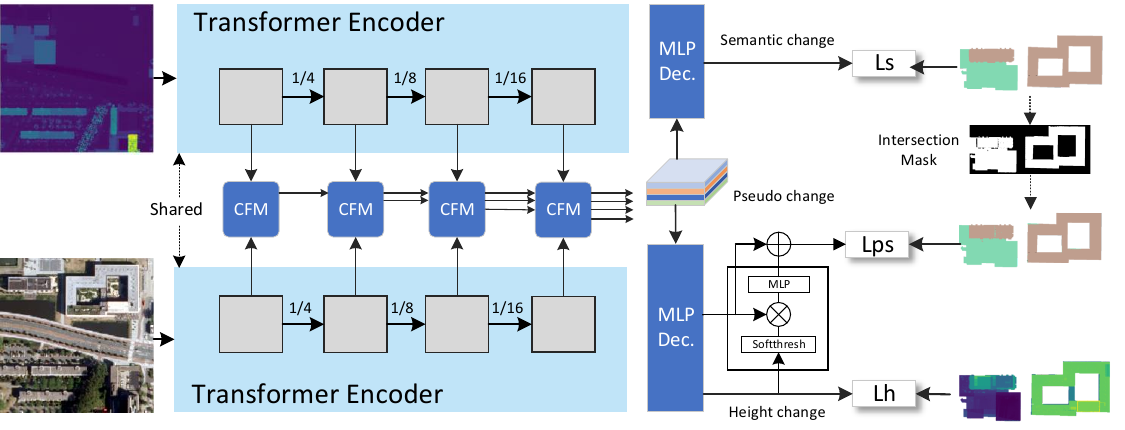}
	\caption{Our Transformer-based multimodal change detection pipeline is named MMCD. It consists of 
		the pyramid backbone with four Transformer layers, the cross-modal fusion module (CFM), and the multi-layer perception (MLP) decoder. The multitask consistency acts as an explicit constraint for enhancing multimodal correlation. }\label{fig_network}
\end{figure*}

\subsection{Problem definition and data preprocessing}
\subsubsection{Multimodal change detection problem}

{Figure \ref{fig:conceptual_pipeline} and \ref{fig_network} depict the pipeline of multimodal change detection problem. We aim to detect both height change and semantic change with pre-event DSM and post-event image. In training phase, the ground truth of height change is obtained by subtracting the multi-temporal DSMs and performing truncated normalization as section \ref{data-prec}. The floating ranged DSM pixels are normalized to grayscale before inputting them to the embedding model. The Transformer-based embedding model $\mathcal{T}$, which is detailed in section \ref{mmcd_backbone}, learns the DSM and image embedding with shared parameters. It can formulated as
	\begin{equation}
		X_H = \mathcal{T}(DSM_{pre}),
		X_I = \mathcal{T}(image_{post}).
	\end{equation}
	Following the backbone, the cross-model fusion process aims to augment the representation of a specific modality by integrating cues from other modalities, as elaborated in Section \ref{mmcd_crossmodal}. Building on the representation consistency between height change and semantic change, we introduce an explicit consistency constraint. This constraint is designed to ensure that both tasks mutually enhance each other's performance, as detailed in Section \ref{mmcd_consistency}.
}

\subsubsection{Data preprocessing}\label{data-prec}

{{\bf Truncated normalization for height changes}: Height change detection leverages the $L2$ regression loss, with practical implementations \cite{marsocci2023inferring, zhao2023semantic} incorporating a Tanh activation layer. This layer normalizes the final decoded layer's output to a range of -1 to 1 for enhanced training stability. Therefore, the ground truth height should be rescaled to $[-1, 1]$ during training. The normalization parameters are obtained by truncating the distribution to include 99.5\% of the height change values from the training set, optimizing the model's focus on dominant height changes. Specifically, the truncated rescaling range is $[-27.29, 87.26]$ (meters) in our training set.
	
	{\bf Gray-scale Normalization for input DSMs}: In practical implementation \cite{zhang2019detecting,sun2021building}, a normalization process is necessary during training image data. For gray-scale RGB images, the integer pixel value ranges from 0 to 255, while the digital surface model could be negative or positive floating values. This causes gradient fluctuation during training and makes it hard to converge \cite{ioffe2015batch}. To this end, we first rescale the height values in DSM into gray-scale as follows:
	\begin{equation}
		Height_{rescaled} = \frac{(Height - min)}{max-min}\times 255
	\end{equation}
	where $min=-10$ and $max=40$. These two hyperparameters are also determined by truncating nearly 99\% of the height value ranges. Then the standard normalization is applied for height values and gray-scale image values. It modifies the data of each channel so that the mean is zero and the standard deviation is one.}

\subsection{MMCD: Transformer-based multimodal change detection network}\label{mmcd_model}

\textbf{Efficient Pyramid backbone.}\label{mmcd_backbone} The Transformer network \cite{vaswani2017attention}, increasingly dominant in recent multimodal data processing, encounters significant challenges when managing bi-temporal inputs that are both high resolution. The architecture, while revolutionary for its attention mechanisms and scalability, struggles with the computational and memory demands posed by large-scale inputs. We aim to develop an efficient and lightweight backbone architecture, considering the typically large data volumes in remote sensing. Therefore, we employ an efficient Transformer block with sequence reduction as our backbone, which is detailed in \cite{wang2021pyramid,bandara2022transformer}. Furthermore, we minimize the embedding dimensions to achieve a more compact model size, effectively halving the complexity compared to ChangeFormer \cite{bandara2022transformer} (see Table \ref{table-semantic}). The pyramid features output from height and image branches are denoted as ${\rm{X}}_{\rm{H}}^n$ and ${\rm{X}}_{\rm{I}}^n$, where $n \in \{1, 2, 3, 4\}.$

\begin{figure*}[!th]%
	\centering
	\includegraphics[width=0.9\textwidth]{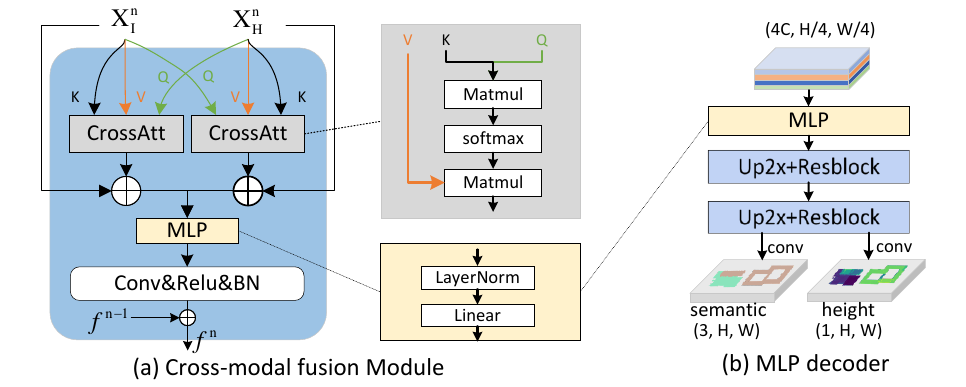}
	\caption{The structure of feature fusion module and decoder in our method. }\label{fig_submodules}
\end{figure*}

\textbf{Cross-modal fusion.}\label{mmcd_crossmodal} As mentioned in \cite{zhu2020edge}, despite the high relevance between depth (or height) data and gray-scale images, establishing the definitive relationship between them is challenging. Inspired by the widely used cross-attention mechanism \cite{chen2021crossvit}, which considers all the multimodal features (such as text, images, audio, or video) as sequences, we compute attention scores by calculating the dot product between the query from one modality and the keys from another modality, followed by a softmax function to normalize the scores. It allows the model to dynamically focus on specific parts of an image based on the context provided by the other modality, and vice versa. The designed cross-modal fusion module is shown in Figure \ref{fig_submodules}(a). It parallelly takes the feature embedding from one modality as query, and the embedding from the other modality as key and value. The standard self-attention component is
\begin{equation}
	\footnotesize
	\operatorname{Attention}({X})\!=\!\operatorname{softmax}(QK)V=\operatorname{softmax}\!\left({{(W_q X)\!(W_k X)}^T}\right) {W_v X},
\end{equation}
where ${W}_q, {W}_k $ and ${W}_v \in {R}^{C \times d}$ are learnable matrices, $X \in \mathbb{R}^{d \times C}$ is input sequence, and the softmax is ${e^{x_i}}/{\sum_{j=1}^{N} e^{x_j}}$. The $C$ is sequence length and $d$ is embedding dimension. The left cross-attention block in Figure \ref{fig_submodules}(a) can be formulated as
\begin{equation}
	\footnotesize
	\operatorname{Cross-attention}({{\rm{X}}_{\rm{H}}^n}, {{\rm{X}}_{\rm{I}}^n})=\operatorname{softmax}\left({{(W_q {\rm{X}}_{\rm{H}}^n) (W_k {\rm{X}}_{\rm{I}}^n)}^T}\right) {W_v {\rm{X}}_{\rm{I}}^n}.
\end{equation}
{This layer allows each modality to query (seek information from) the other modality's representations. The attention scores $\left({{(W_q {\rm{X}}_{\rm{H}}^n) (W_k {\rm{X}}_{\rm{I}}^n)}^T}\right)$ are used to create a weighted sum of the value vectors from the attended modality, allowing the model to focus more on the relevant features. This results in a richer, contextually informed representation that combines information from both modalities. The multimodal feature spaces before and after the cross-attention is depicted in Figure \ref{fig:vis_tsne3}.} The CFM uses two symmetrically arranged cross-attention operators for capturing mutual relationships between features derived from the DSM and image branches. Next, they are merged through MLP and {the} convolutional unit, which is then pixel-wise added to the previous feature layer $f^{n-1}$ to obtain $f^n$.

\textbf{MLP decoder.}\label{mmcd_decoder} Figure \ref{fig_submodules}(b) depicts the streamlined structure of our MLP decoder. We use non-parameterized up-sampling, while ChangeFormer \cite{bandara2022transformer} utilizes learnable transposed-convolution that incurs higher computation cost. Furthermore, the residual convolutional blocks \cite{he2016deep} are utilized to enhance the local relations during the up-sampling process. The final semantic and height change maps are {generated} through $3\times3$ convolution block {following the approach outlined in} MTBIT \cite{marsocci2023inferring}.

\begin{figure*}[!h]%
	\centering
	\includegraphics[width=1.0\textwidth]{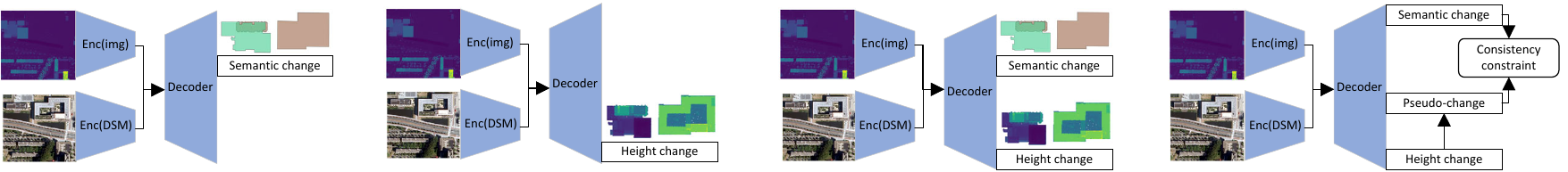}
	\caption{The network variations of our method, arranged from left to right, include: the only semantic change detection branch, only height change detection branch, the multitask branch, and the multitask branch with consistency constraint.}\label{variations}
\end{figure*}

{\textbf{Model variations.} 
	Figure \ref{fig_network} depicts the final model architecture, while the variations of it, such as "only semantic cd branch", "only height cd branch", and "semantic + height cd branch" in Table \ref{table-ablation_semantic}, are briefly depicted Figure \ref{variations}. }

\subsection{Multitask consistency by predicting the pseudo semantic change}\label{mmcd_consistency}

\begin{figure*}[!th]%
	\centering
	\includegraphics[width=1.0\textwidth]{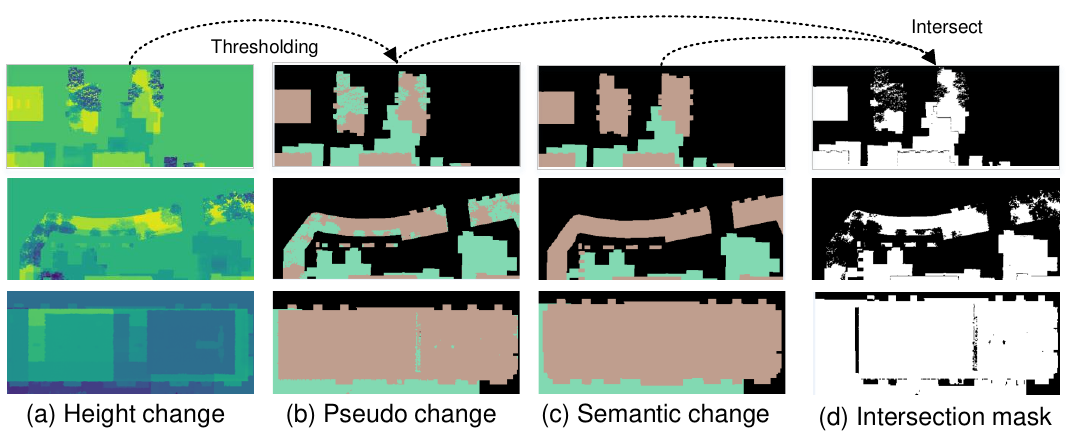}
	\caption{The inconsistency of multimodal change labels. (a) height change; (b) pseudo change by classifying the zero height as unchanged, positive height as newly-built and negative height as demolished regions; (c) semantic change; (d) intersection mask between height change and pseudo change, where the intersection rate of training, validation, and testing set are 79.71\%, 89.23\% and 90.03\% respectively.}\label{fig_pschange}
\end{figure*}

{We bridge the gap between height change detection and semantic change detection tasks by imposing an explicit consistency constraint, facilitated through the prediction of an auxiliary pseudo-change map. Our approach addresses the inherent discrepancy between semantic and height changes—the former being categorical with discrete values, and the latter represented by continuous floating-point values.}
As shown in Figure \ref{fig_pschange}(a)(b), by adopting zero as a threshold for height change, we obtain a classification map termed pseudo change that includes three classes: 0 (the background), 1 (positive height change) and the -1 (negative height change), which differs from the semantic change (Figure \ref{fig_pschange}(c)). The hard thresholding can be formulated as 
\begin{equation}
	T_h(x)=\left\{
	\begin{aligned}
		1&, x > 0  \\[-5pt]
		0&, x = 0 \\[-5pt]
		-1&, x < 0.
	\end{aligned}
	\right.
\end{equation}
Since it is not differential, we adopt a soft thresholding function as follow
\begin{equation}
	\label{eq_softthresh}
	T_s(x)=
	\begin{aligned}
		2\times{sigmoid(\frac{x}{t})}-1,
	\end{aligned}
\end{equation}
where $sigmoid=1/(1+e^{-x})$, and $t$ is a positive temperature parameter for controlling the sharpness of the transition around zero. In our experiment, we set $t=0.5$. Smaller $t$ leads to a more accurate approximation to $T_h(x)$. This can be implemented with a sigmoid and an MLP layer for introducing strong prior to the pseudo-change branch. The pseudo change highly overlaps with semantic change but is not totally the same as shown in Figure \ref{fig_pschange}(d). Therefore, only overlapped areas are considered when measuring the consistency between them. The objective is to minimize the following objective function
\begin{equation}
	\footnotesize
	\mathcal{L}_{\text {consistency }}=\min _{\text {Pred}_{sc}, \text {Pred}_{psc}}
	\begin{aligned}
		(GT_{psc}\cap GT_{sc})*|Pred_{psc}-Pred_{sc}|,
	\end{aligned}
\end{equation}
where $|\cdot|$ is kind of distance. $GT_{sc}$ and $GT_{psc}$ are ground truth of semantic and pseudo change. $Pred_{sc}$ and $Pred_{psc}$ are corresponding model prediction. In practical implementation, the semantic and pseudo change branches are separately supervised with $GT_{sc}$ and $GT_{psc}$ respectively, which leads to consistency in their overlapping regions.

\subsection{Loss function}

We employ the weighted cross-entropy loss for both the semantic and pseudo change detection branches and utilize mean-square error ($L2$ loss) for the height change detection branch as \cite{marsocci2023inferring}, which are denoted as $\mathcal{L}_{\text height}$, $\mathcal{L}_{\text {pseudo}}$, and $\mathcal{L}_{\text {semantic}}$ respectively. The final training loss is 
\begin{equation}
	\mathcal{L}_{\text {total}}=\lambda_1 \cdot \mathcal{L}_{\text {pseudo}}+\lambda_2 \cdot \mathcal{L}_{\text {height}}+\lambda_3 \cdot \mathcal{L}_{\text {semantic}}
\end{equation}
where $\lambda_1$, $\lambda_2$, and $\lambda_3$ are fixed loss weights, which are 0.2, 0.2, and 0.6 in our experiment setting.

\section{Hi-BCD: A multimodal dataset for building change detection between height map and optical image}\label{sec_data}

\begin{table}[!tp]
	\caption{The main details of typical existing change detection datasets. First three rows: single-modal datasets. Last four rows: multimodal datasets. Note that the 3DCD \cite{marsocci2023inferring} employs bi-temporal images to infer changes across multiple modalities.}
	\label{table-data_compare}
	\vspace{2mm}
	\centering
	\scalebox{0.85}{
		\begin{tabular}{cccccc}
			\toprule \text {Name} & \text {N. images} & \text {Tile size} & \text {Resolution} & \text {CD map} & \text {Classes} \\
			\midrule
			\text{LEVIR-CD} \cite{chen2020spatial} & 637 & $1024\times1024$ & 0.5 m & 2D & 2\\
			\text{Hi-UCD} \cite{tian2022large} & 40800 &  $512\times512$& 0.1 m & 2D & 9 \\
			\text{DynamicEarthNet} \cite{toker2022dynamicearthnet} & 54750& $1024\times1024$& 3 m & 2D & 7\\
			\hline 
			\text{Shuguang} \cite{chen2022unsupervised} & 1 & $921\times593$ & - & 2D & 1 \\
			\text {multimodalCD \cite{zhang2019change}} & 3615 & $100 \times 100$ & 0.1 m & 2D & 2 \\
			\text {3DCD \cite{marsocci2023inferring}} & 472 & $400 \times 400$ & 0.5 / 1m & 2D / 3D & 1 \\
			\text {Hi-BCD (ours)} & 1500 & $1000 \times 1000$ & 0.25 m & 2D / 3D & 2 \\
			\bottomrule
	\end{tabular}}
\end{table}

{\bf Existing datasets and limitations}. Table \ref{table-data_compare} presents a concise comparison of existing change detection datasets, encompassing both single-modal and multimodal datasets. A substantial volume of bi-temporal 2D image datasets available, supporting high-resolution, high-frequency, and multi-class change analysis. Despite the abundance of single-modal datasets, limited research has ventured beyond 2D change detection using multimodal data. The {\bf Shuguang} dataset used in \cite{chen2022unsupervised} contains a pair of SAR and optical images for detecting 2D construction change. The constraint of a small sample size dictates that {most} earlier methods can only be developed from an unsupervised standpoint. {The} {\bf multimodalCD} \cite{zhang2019change} {dataset} incorporates multi-view image-based and ALS-based point clouds, which are transformed into DSMs to detect binary 2D changes exclusively. However, it confines its focus to 2D changes within small tile sizes, even though it contains beyond-2D information.
{The} {\bf 3DCD} \cite{marsocci2023inferring} {dataset} employs bi-temporal images to identify not only binary building changes but also variations in height. However, the resolution of height change is merely half that of 2D changes, and it exclusively classifies binary changes.
{To establish a new foundation for change detection, we introduce the Hi-BCD dataset. It leverages multimodal inputs to simultaneously output semantic and height changes, providing generously sized tiles and high-resolution 2D and 3D semantic change maps.}

\begin{figure*}[!th]%
	\centering
	\includegraphics[width=1.0\textwidth]{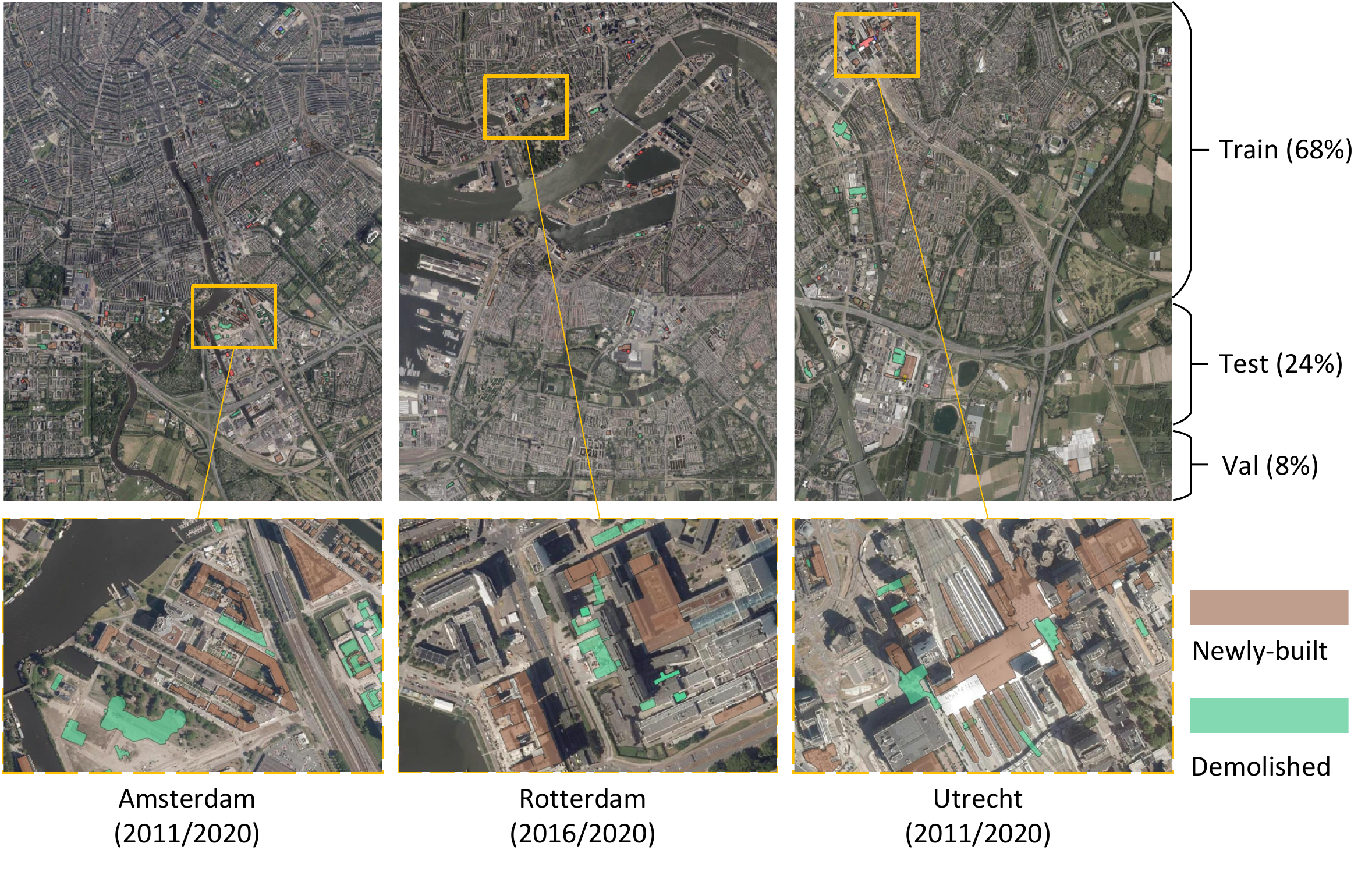}
	\caption{Hi-BCD dataset. It encompasses three cities in the Netherlands and provides two types of building changes. The dates of early and late periods are denoted under each city. For each city, 68\%, 24\%, and 8\% of the tile are used for training, 
		testing, and validation, respectively.}\label{fig_dataset}
\end{figure*}

\begin{figure*}[!th]%
	\centering
	\includegraphics[width=1.0\textwidth]{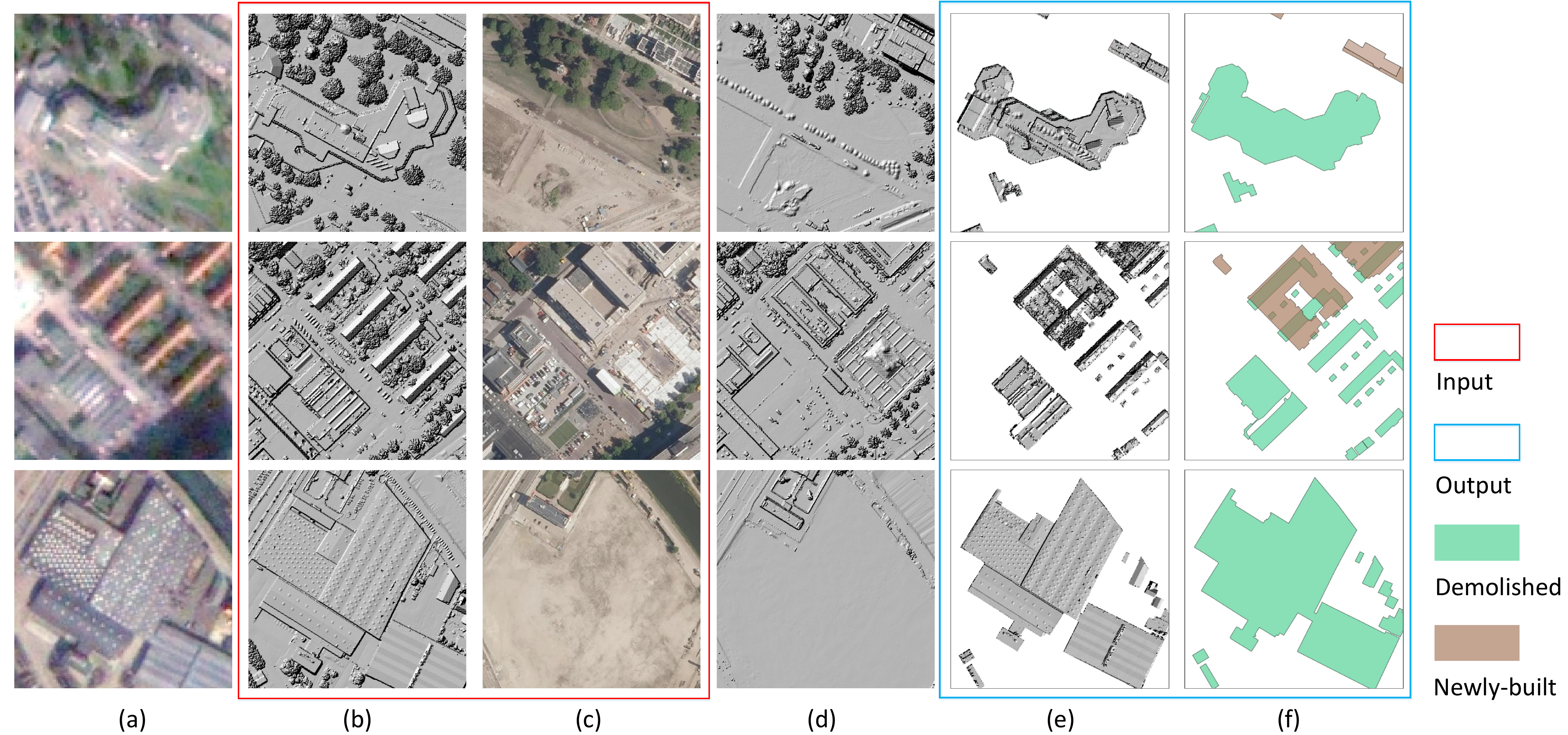}
	\caption{Examples of Hi-BCD dataset, where the DSMs are displayed in hillshade manner (A widely used visual technique to give a three-dimensional appearance from DSM data). The (a) and (b) are pre-temporal images (\textbf{2.0m} resolution) and DSMs (\textbf{0.25m} resolution). The (c) and (d) are post-temporal images and DSMs (all \textbf{0.25m} resolution). The (e) and (f) are height and semantic changes between the multimodal temporal inputs.}\label{fig_dataset_sample}
\end{figure*}

\begin{figure*}[!bh]%
	\centering
	\includegraphics[width=1.0\textwidth]{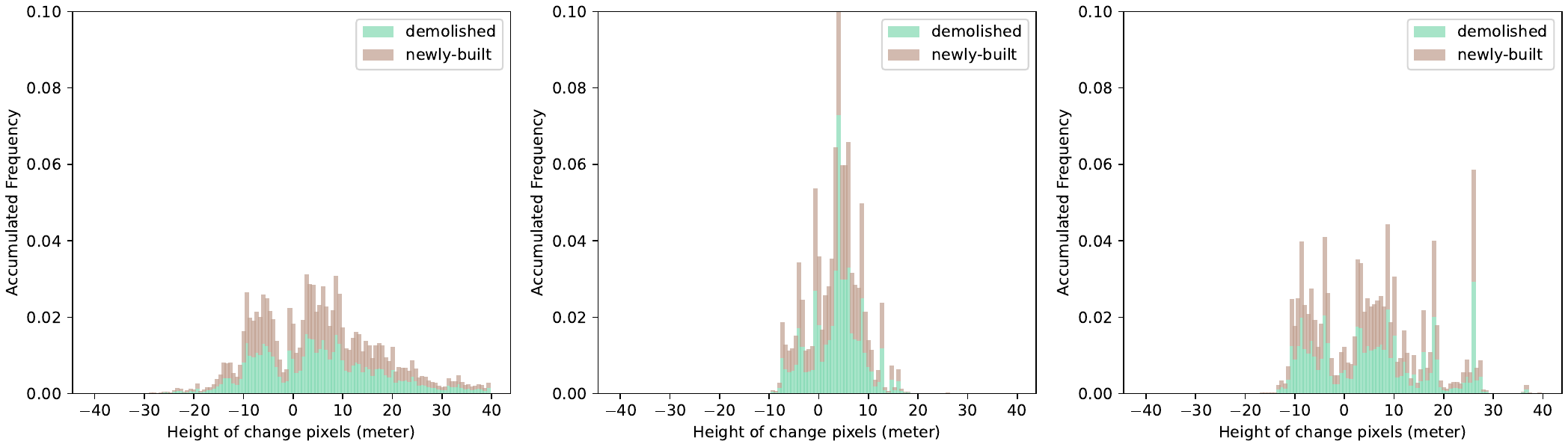}
	\caption{The cumulative frequency of height values for two types of change pixels.}\label{fig_dataset_dist}
\end{figure*}

\textbf{Dataset overview.} As depicted in Figure \ref{fig_dataset}, our study area involves three cities in the Netherlands, including Amsterdam, Rotterdam, and Utrecht. The dates of the pre-temporal and post-temporal periods are indicated beneath each city in Figure \ref{fig_dataset}. We can observe significant variation in the capture dates of the pre-temporal DSM, which reflects the extensive updating period of high-dimensional data. This hinders the application of high-dimensional change detection with dual-temporal 3D data. {The DSM data is generated from point clouds using grid sampling and strictly orthogonal projections, whereas the aerial images are ortho-photographs.}
{The data volume for each city comprises five hundred of} $1000 \times 1000$ DSM-to-image pairs with a ground sampling distance of 0.25 meter (Figure \ref{fig_dataset_sample}(b)(c)). The corresponding multi-class 3D and 2D changes are shown in Figure \ref{fig_dataset_sample}(e)(f). The vertical accuracy is about 0.15 meters \cite{cserep2023distributed}. For each city tile, 68\%, 24\%, and 8\% are allocated for training, testing, and validation respectively. Two types of change including 'newly-built' and 'demolished' are defined in the dataset. 
More details about change objects, pixels, and samples are provided in Table \ref{table-data}. Figure \ref{fig_dataset_dist} portrays the cumulative frequency of height for the two types of change. 

\begin{table}[tbp!]
	\caption{The main details of the Hi-BCD dataset, including changed objects, pixels, and sample amount of three cities in the Netherlands.}
	\label{table-data}
	\vspace{2mm}
	\centering
	\scalebox{0.85}{
		\begin{tabular}{ccccc}
			\toprule
			Attribute & Category & Amsterdam  & Rotterdam  & Utrecht  \\
			\midrule
			\multirow{2}*{\makecell[c]{changed \\ objects}} 
			& newly-built & 389 & 510 & 458\\
			\addlinespace[-0.5ex]
			~ & demolished & 251 & 229 & 187\\
			\cmidrule(r){2-5}
			\multirow{2}*{\makecell[c]{changed \\ pixels}} & amount & 6.625M & 5.139M & 7.73M \\
			\addlinespace[-0.5ex]
			~ & $prop_{/total}$ &  1.3\%&1.0\%&1.5\% \\
			\addlinespace[-0.5ex]
			\cmidrule(r){2-5}
			\multirow{2}*{\makecell[c]{samples \\ (size: $1k\times1k$ )}} & total & 500 & 500&500\\
			\addlinespace[-0.5ex]
			~ & with change & 40.8\%  & 34.2\% & 43\%  \\
			\bottomrule
	\end{tabular}}
\end{table}

{\bf Annatation procedure.} We build the dataset based on AHN\footnote{http://www.ahn.nl/} (Actueel Hoogtebestand Nederland), the nationwide elevation data project in the Netherlands. Specifically, the early-period elevation data {is obtained by rasterizing} point clouds from AHN3 (2011-2019), while the aerial images\footnote{https://www.beeldmateriaal.nl/over-beeldmateriaal} with the close date to AHN4 (2020-2022) are used as late period. The annotation procedure is as follows:

1) {\bf Change definition.} The construction and demolition of buildings are annotated based on the difference map between AHN3 and AHN4. Since the capture date of aerial images does not precisely align with that of the point cloud, we have focused our annotations on multi-class building changes that remain relatively stable within a one-year time frame, while excluding highly dynamic changes such as trees and vehicles. Note that the demolished and newly-built buildings do not highly correspond to the negative and positive height change values respectively, due to tree occlusions and the penetration of low-reflectivity surfaces such as glass roofs, as depicted in Figure \ref{fig_pschange}. We define a mask that indicates whether the building changes are highly relevant to their height change values, which is 
\begin{equation}
	M(i,j) = \left\{
	\begin{aligned}
		1&, \text{highly relevant}\\[-3pt]
		-1&, \text{otherwise}
	\end{aligned}
	\right.
\end{equation}
2) {\bf Edge situation.} The second row in column (f) of Figure \ref{fig_dataset_sample} illustrates a complex scenario where various building changes intersect. This suggests a sequence of events where a building is first demolished and subsequently replaced by a new one. 
For such a situation, the change type is determined based on the elevation difference, where pixels with a positive change in height are categorized as newly built, while pixels with a negative change in height are associated with demolished building changes. While we have provided a simplified representation by aggregating these overlapping changes into single-type changes, there remains an opportunity for future research to delve into finer sub-situations, describing the entire evolution process.

3) {\bf Change label generation.} Based on the above-mentioned change definition and edge situation, the change map (CM) is formulated as 
\begin{equation}
	\footnotesize
	CM(i,j) = \left\{
	\begin{aligned}
		\text{demolished}&, \Delta H \cdot M(i,j)< 0 \text{ and building in early period} \\[-3pt]
		\text{newly-built}&, \Delta H \cdot M(i,j)> 0 \text{ and building in late period} \\[-3pt]
		\text{background}&, otherwise.
	\end{aligned}
	\right.
\end{equation}
where $\Delta$H is the elevation difference between the bi-temporal DSM data, i.e., 
$\Delta H =\text{DSM}_{AHN4}-\text{DSM}_{AHN3}$. The definition of the first two cases implies a sub-situation of bi-temporal buildings. It may be a little misaligned with the building in image data due to viewpoint distortion. The labels for 3D height changes correspond to the masked regions in the AHN4-to-AHN3 difference map.

4) {\bf Tile splitting.} The original tile size of each city is 25000$\times$20000 with a pixel granularity of 0.25 meter, which is subsequently split into 500 pairs of 1000$\times$1000 sized samples. Note that we retained samples that do not include changes to better reflect the data distribution of the real-world scenario.

{\bf Challenges of Hi-BCD dataset}. The fundamental challenge of our dataset is to learn the representation of multi-class elevation changes from multimodal and cross-dimensional inputs. There are some inherent misalignments between the multimodal inputs. 1) The pixels hold diverse implications as the DSMs rasterize the height dimension from Lidar point clouds and represent the absolute land elevation, while the images reflect the intensity of visible light. They exhibit vastly different numerical ranges, where the height range is $[-8.24,183.64]$ for {the} original DSM and $[-99.55,134.21]$ for the changes in the training set, while the image values are in grayscale $[0,255]$. 2) Their distribution differs a lot as the DSMs exhibit similar height in the ground regions, while the images portray different colors and textures for various land covers. 3) There exists geometry misalignment due to viewpoint distortion of aerial images although they utilize the same coordinate system.
Furthermore, severe change-unchange imbalance can be observed in Table \ref{table-data}.
Additionally, there is inconsistency between the semantic change labels and the height change labels as shown in Figure \ref{fig_pschange}.

\section{Experiments and Results}
\subsection{Experiment setting}
\textbf{Implementation details.}
The Tanh function normalizes height outputs to $[-1,1]$. The elevation scale of training set $[-27.29,87.26]$ covering 99.5\% of pixels is used for denormalization. We set class weights of 0.05, 0.95{,} and 0.95 to the background, demolished, and newly-built areas for weighted cross-entropy loss. All the models are pre-trained in LEVIR-CD \cite{chen2020spatial} and then trained for 300 epochs with equivalent batch size of 8. At ease of multi-scale downsampling, the original tile size of 1000$\times$1000 is adjusted to 1024$\times$1024 during training. More details can be found here\footnote{https://github.com/qaz670756/MMCD}.

\textbf{Metrics}. For semantic change detection evaluation, we used the mean intersection over union (mIoU) and F1-score as denoted in \cite{marsocci2023inferring}. For the height change detection, we keep consistent with {relevant} research \cite{marsocci2023inferring,zhao2023semantic,amirkolaee2019height}, including the following metrics:
\begin{itemize}[itemsep=0pt]
	\item Root Mean Square Error (RMSE): {\footnotesize$\sqrt{\frac{1}{n} \sum\left(H_r-H_e\right)^2}$}
	\item Mean Average Error (MAE): {\footnotesize$\frac{1}{n} \sum {\left|H_r-H_e\right|}$}
	\item Root Mean Square Error (cRMSE): {\footnotesize$\sqrt{\frac{1}{n} \sum\left(H_r-H_e\right)^2}$}
	\item Average relative error (cRel): {\footnotesize$\frac{1}{n} \sum \frac{\left|H_r-H_e\right|}{H_r}$}
	
	\item Mean normalized cross correlation (ZNCC): {\footnotesize$\frac{1}{N} \sum_i^N \frac{\left(H_{r i}-\mu_{H_r}\right)\left(H_{e i}-\mu_{H_e}\right)}{\sigma^{H_r} \sigma^{H_e}}$}
\end{itemize}

where $H_r$ denotes the reference height, $H_e$ denotes the estimated
height, and $N$ denotes the estimated pixel count. $\mu$ and $\sigma$ are the mean values and standard deviations of $H_r$ and $H_e$, respectively. The $\text{cRMSE}$ and $\text{cRel}$ {indicate} that only changed areas are considered. The ZNCC quantifies the spatial correlation between output and ground truth, while the other metrics measure the degree of absolute errors at each pixel in meters. Moreover, we {include} the million parameter count (MParams) and Giga Floating-Point Operations (GFLOPs) {as metrics} to compare the model complexity \cite{chen2021remote,liu2021zoominnet}. 

\textbf{Compared methods}. Limited research, such as \cite{marsocci2023inferring}, {has concurrently} detected semantic and height changes. To provide a benchmarking, we follow the structure of MTBIT \cite{marsocci2023inferring} that maintains the original change detection network while slightly modifying the decoder with an additional height change detection branch. Among the selected methods, the {FC-Siamese} \cite{daudt2018fully} is the first fully convolution-based change detection architecture widely used for comparison. The {SNUNet} \cite{fang2021snunet} is a state-of-the-art CNN-based method {featuring a} densely connected backbone. The {ChangeFormer} \cite{bandara2022transformer} is a Transformer-based method that yields promising results in most change detection benchmarks. The {P2VNet} \cite{lin2022transition} models the change process in a novel multi-frame transition perspective. The {MTBIT} extends the Transformer-based BIT \cite{chen2021remote} for simultaneously detecting semantic and height changes.

\subsection{Comparison with state-of-the-arts}\label{sec_compare}

\begin{table*}[!th]
	\caption{The semantic change detection performance before and after attaching the height predicting branch. The methods with $*$ are originally designed with a height branch. The colors {\color{red}red}, {\color{green}green}, and {\color{blue}blue} indicate the top three results.}
	\label{table-semantic}
	\vspace{2mm}
	\centering
	\scalebox{1.0}{
		\begin{tabular}{cc|cccc|cccc|cc}
			\toprule
			\multirow{2}*{Method} & \multirow{2}*{Year}&\multicolumn{4}{c|}{only semantic cd branch}&\multicolumn{4}{c|}{semantic + height cd branch} &\multicolumn{2}{c}{Complexity (two branches)}\\
			\cline{3-12}
			~ & ~ & $IoU_{D}$ $\uparrow$& $IoU_{N}$$\uparrow$& mIoU$\uparrow$& F1-score$\uparrow$ &  $IoU_{D}$ $\uparrow$& $IoU_{N}$$\uparrow$& mIoU$\uparrow$& F1-score$\uparrow$& MParams$\downarrow$ & GFLOPs$\downarrow$   \\
			\hline
			FC-Siamese \cite{daudt2018fully} & 2018 &27.77&29.18&28.48&44.33& 27.60&28.34&27.97&43.72&{1.552}&{92.908} \\  
			SNUNet \cite{fang2021snunet} & 2021 &25.47 &26.07&25.77 & 40.98& 20.77&22.97&21.87&35.87& {3.012}& 220.696\\
			ChangeFormer \cite{bandara2022transformer} & 2022&{\color{red}47.17}&{\color{green}36.67}& {\color{red}41.92}&{\color{red}58.88}& {\color{green}38.89}&{\color{red}41.45}&{\color{green}40.17}&{\color{green}57.31} & 29.75&340.165 \\ 
			P2VNet \cite{lin2022transition}&2022&37.41&{\color{blue}30.61}&{\color{blue}34.00}&{\color{blue}50.65}&{\color{blue}35.63}&{\color{blue}30.46}&{\color{blue}33.04}&{\color{blue}49.62}& 5.425&527.442\\
			*MTBIT \cite{marsocci2023inferring} & 2023 &{\color{blue}37.44}&29.97&33.71&50.31 &
			34.40&27.04&30.72&46.88&{15.2}&{154.72} \\
			\hline
			*Ours & 2023 & {\color{green}43.76}&{\color{red}39.10}&{\color{green}41.43}&{\color{green}58.55} & {\color{red}44.29}&{\color{green}40.90}&{\color{red}42.59}&{\color{red}59.72} & 11.659& {168.893}\\
			\bottomrule
	\end{tabular}}
\end{table*}

In this section, we evaluate semantic and height change detection to explore how these two tasks influence each other. Note that semantic change detection refers to multi-class 2D change detection in our context. The model operates in a multitask situation when performing joint semantic and height change detection. Otherwise, it is in a single-task setting {when addressing only one of these tasks}.

\textbf{Semantic change detection}.   
In Table \ref{table-semantic}, our method achieves competitive semantic results {across} both single-task and multitask settings. Specifically, by employing a single semantic change detection branch, we achieve close results to ChangeFormer with only half the model complexity. Besides that, the model with the largest number of parameters (ChangeFormer) and highest computational cost (P2VNet) achieved the second and third best results, respectively. 
When augmented with our consistency-enhanced height prediction branch, the metric numbers exhibit continued improvement. On the contrary, the other methods that attached to the height change detection branch without consistency constraint, show a notable degradation. {This suggests} that the added height change detection branch hinders the learning of the semantic branch, where a similar phenomenon is also observed in prior works \cite{zhu2020edge,liu2021conflict}. Different optimization objectives among multiple tasks can lead to potential mutual interference during feature optimization. With the help of the consistency constraint, our method prevents performance degradation due to interference from the height change detection branch.

\begin{table*}[!th]
	\caption{The height change detection performance without and with semantic branch. The {underlined} \underline{numbers} indicate a decrease with the attached semantic change detection branch compared to its single-branch counterpart. The methods with $*$ are originally designed with a height branch. The colors {\color{red}red}, {\color{green}green} and {\color{blue}blue} indicate the top three.}
	\label{table-height}
	\vspace{2mm}
	\centering
	\scalebox{1.0}{
		\begin{tabular}{c|ccccc|ccccc}
			\toprule
			\multirow{2}*{Method} & \multicolumn{5}{c|}{only height cd branch} & \multicolumn{5}{c}{semantic + height cd branch}\\
			
			\cline{2-11}
			
			~ & RMSE$\downarrow$ & MAE$\downarrow$ & cRMSE$\downarrow$ & cRel$\downarrow$& cZNCC$\uparrow$ &  RMSE$\downarrow$ & MAE$\downarrow$ & cRMSE$\downarrow$ & cRel$\downarrow$& cZNCC$\uparrow$ \\
			\hline
			FC-Siamese   &{\color{blue}1.505}&0.446&{\color{blue}8.622}&{\color{green}1.838}&{\color{blue}0.274}&1.461&{\color{blue}0.309}&\underline{8.995}&{\color{green}1.506}&{0.373} \\
			SNUNet  &1.574&0.498&{9.397}&2.988&0.186&\underline{1.671}&\underline{0.779}&9.216&2.704&0.265 \\ 
			ChangeFormer  &1.658&{\color{red}0.313}&{\color{green}8.404}&{2.110}&{\color{green}0.308}&{\color{green}1.343}&\underline{0.402}&{\color{red}8.204}&\underline{2.485}&{\color{green}0.394}\\
			P2V-CD&{\color{green}1.392}&{\color{blue}0.399}&9.037&{\color{blue}1.937}&0.261&{\color{blue}\underline{1.408}}&{\color{green}{0.305}}&{8.932}&{\color{red}1.463}&{\color{blue}0.377}\\
			*MTBIT&  {1.530}&0.475&9.441&{\color{red}1.780}&0.179&1.457&0.400&{\color{blue}8.563}&\underline{1.987}&0.345\\
			\hline
			*Ours& {\color{red}1.273}&{\color{green}0.397}&{\color{red}8.317}&2.711&{\color{red}0.379}&{\color{red}1.267}&{\color{red}0.290}&{\color{green}8.281}&{\color{blue}{1.900}}&{\color{red}{0.394}}\\
			\bottomrule
			
	\end{tabular}}
\end{table*}

\begin{figure*}[!th]%
	\centering
	\includegraphics[width=1.0\textwidth]{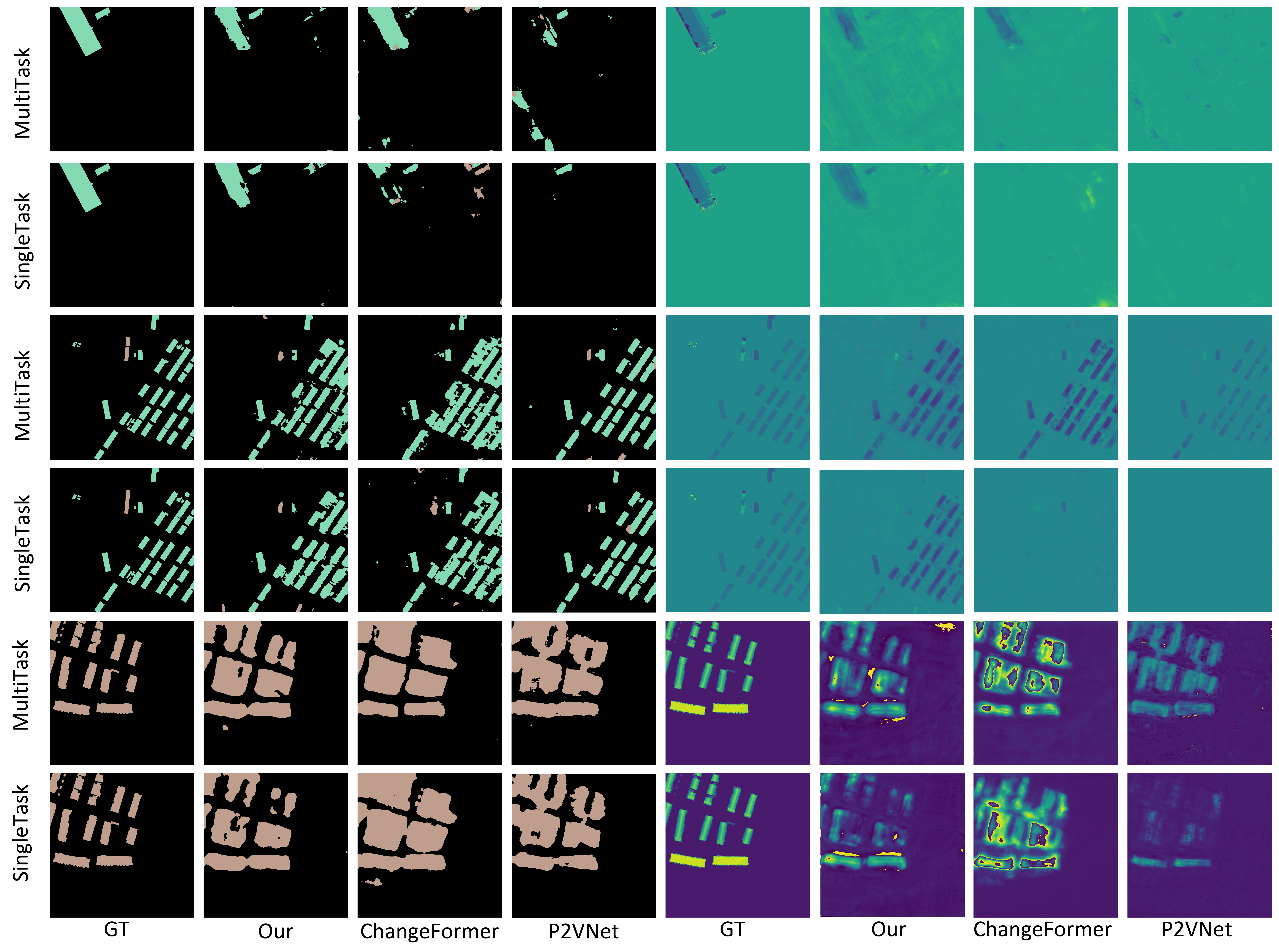}
	\caption{Visual comparison of semantic (first four columns) and height (last four columns) changes for the top-three methods in single-task (only semantic or height branch) and multitask change detection settings. Note that height changes of our model are from the consistency augmented height branch corresponding to row 3 of Table \ref{table-ablation_semantic}.}\label{fig_singlevsdouble}
\end{figure*}

\textbf{Height change detection.} {Table \ref{table-height} reveals an intriguing pattern:} Among the 30 metric results obtained with a height change detection branch, 23 of them gain improvements when working with a semantic branch, while only 7 of them show a decrease. This phenomenon {underscores} the positive impact of learning the shared representation through the implicit hints from semantic change. As also denoted in \cite{zhu2020edge}, the object boundaries are easier to capture from the semantic map compared to the depth map. {Incorporating a dedicated branch for predicting pseudo changes from the height map establishes a clear correlation between semantic and height changes.} This enhancement is evident in the improved performance of height change detection across five different metrics. {With only half the model complexity of ChangeFormer, our model reaches the top in most metrics. 
	Note that this enhancement, facilitated by a consistency constraint, can be conveniently adapted to other methods, yielding promising improvements as showcased in Table \ref{table-mc_for_sotas}.}

\begin{figure*}[!t]%
	\centering
	\includegraphics[width=1.0\textwidth]{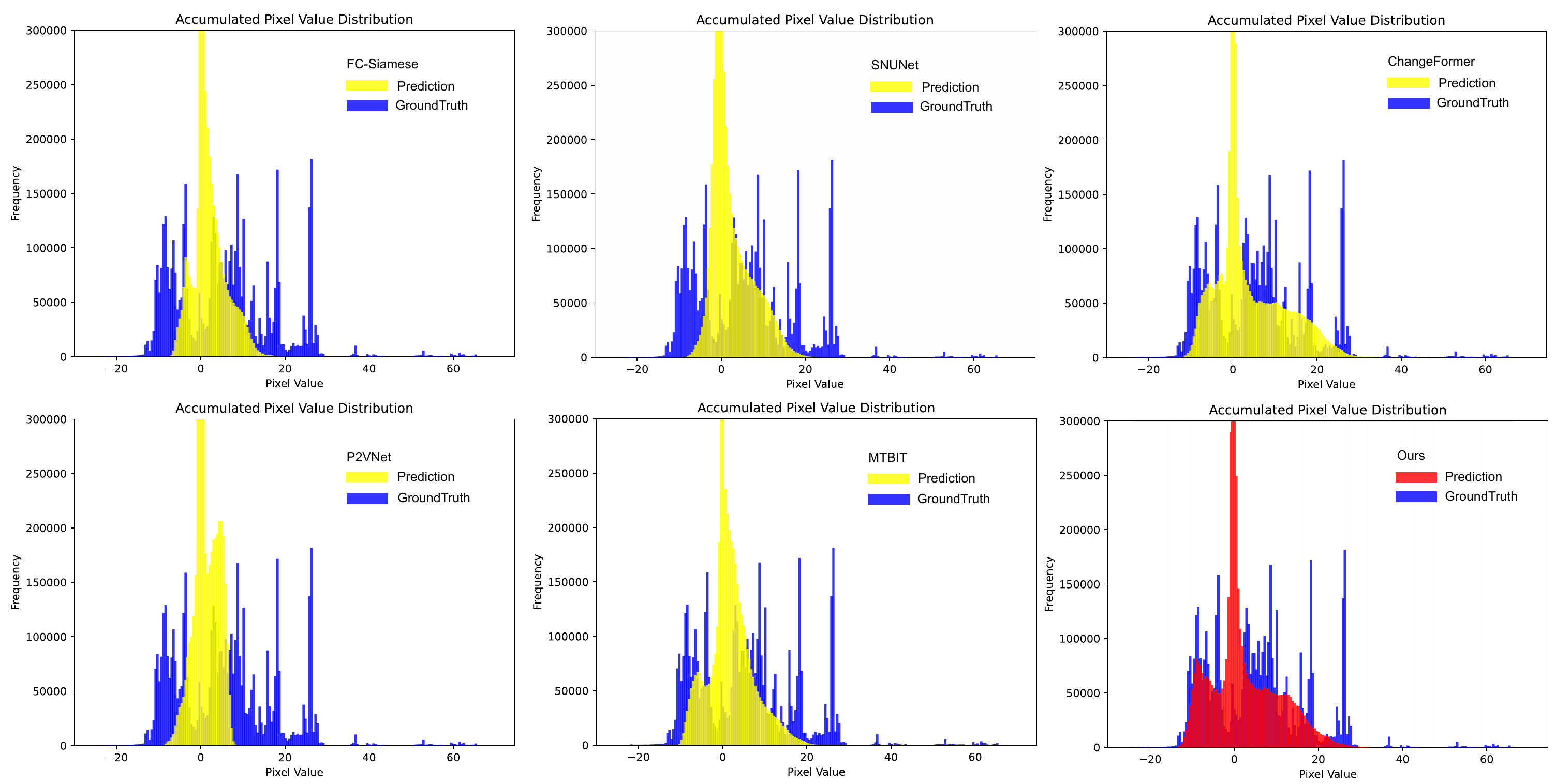}
	\caption{The comparison of predicted height distribution, where blue denotes ground truth, yellow and red denote the other methods and ours. }\label{fig_heightbars}
\end{figure*}

\textbf{Qualitative Results}. 
Figure \ref{fig_singlevsdouble} depicts the visual comparison of semantic and height change detection for the top three methods. The tendency could be observed in the first four columns of Figure \ref{fig_singlevsdouble}. {In} the last four columns, collapsed results {are evident} for models {lacking} semantic hints in single-task setting, {with ours being the exception}. Note that the single-task visual results of our method are derived from the height change detection branch that is augmented with multitask consistency.

\begin{figure*}[!ht]
	\centering
	\includegraphics[width=1.0\textwidth]{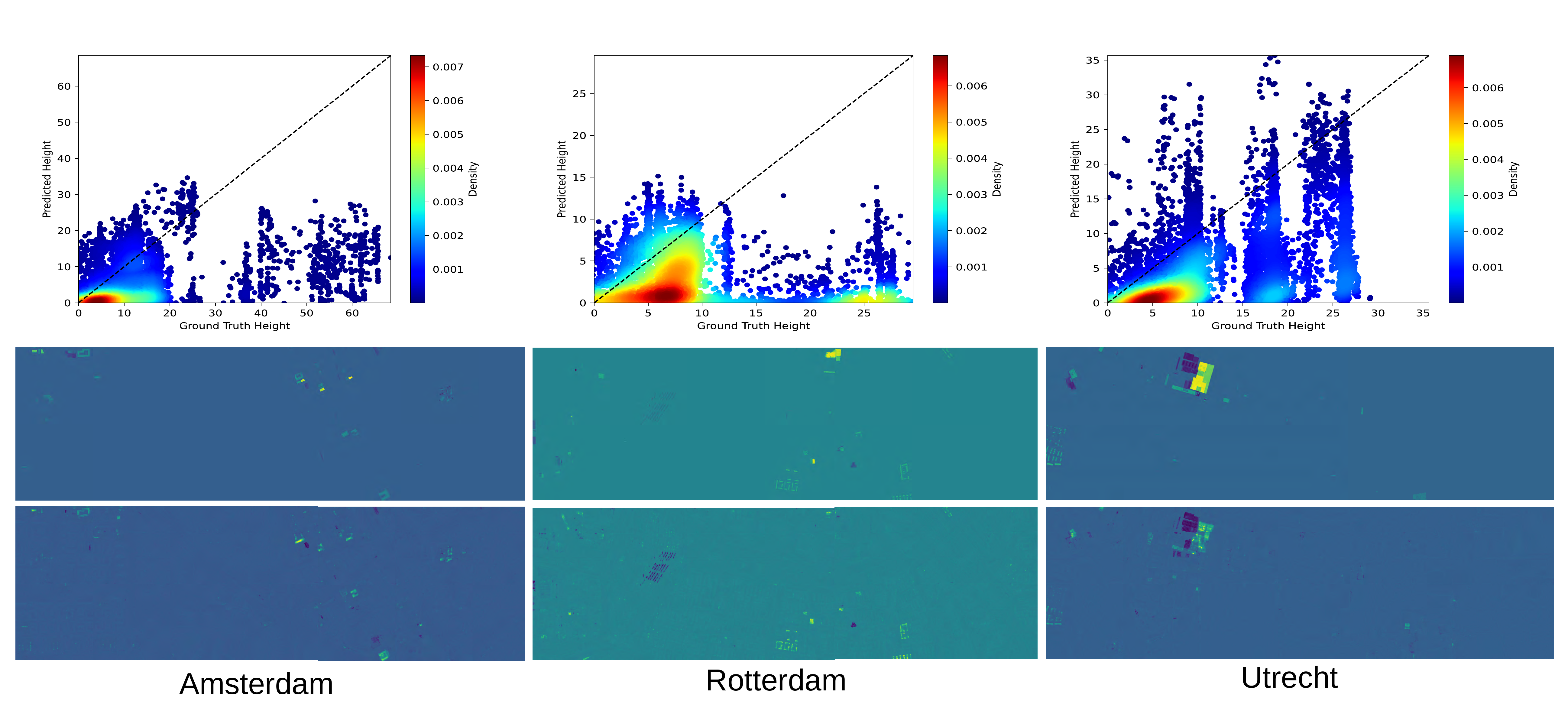}
	\caption{Large scale visualization of height change predictions in three cities (please enlarge for more details). First row: the scatter between ground truth and predicted heights with density coloring map. Second row: ground truth of the whole test set. Third row: the predictions of our model.}
	\label{fig:large_3d}
\end{figure*}

\begin{figure*}[!h]
	\centering
	\includegraphics[width=1.0\textwidth]{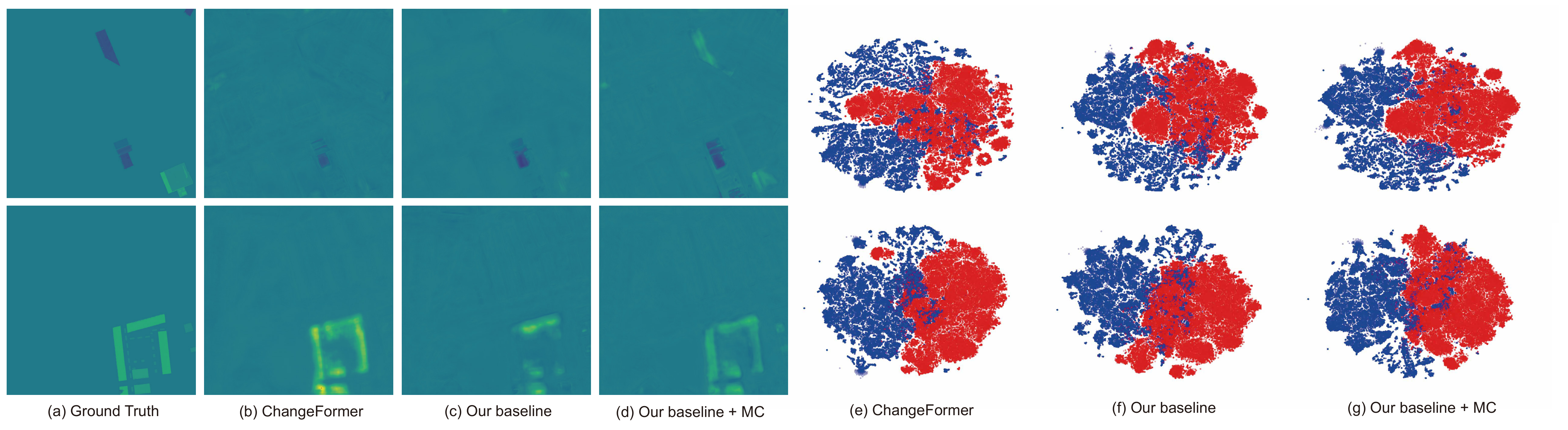}
	\caption{The feature space of the DSM and image branches visualized by t-SNE. The red indicates an image feature point and the blue indicates a DSM feature point.}
	\label{fig:vis_tsne1}
\end{figure*}

Figure \ref{fig_heightbars} presents a comparative analysis of height prediction, illustrating the distribution patterns across various methods. In terms of the height change value range, other change detection methods generally tend to underestimate, whereas our method delivers a more accurate range. Regarding the overall distribution, most methods exhibit a single peak around zero, except for ChangeFormer, MTBIT, and our approach, which align more closely with the actual distribution that features at least two significant clusters. However, for all methods, the dominant portion of height outputs clusters near zero. This reflects the background to changed areas imbalance. Addressing this problem remains a direction for future research.

\begin{figure*}[!t]
	\centering
	\includegraphics[width=1.0\textwidth]{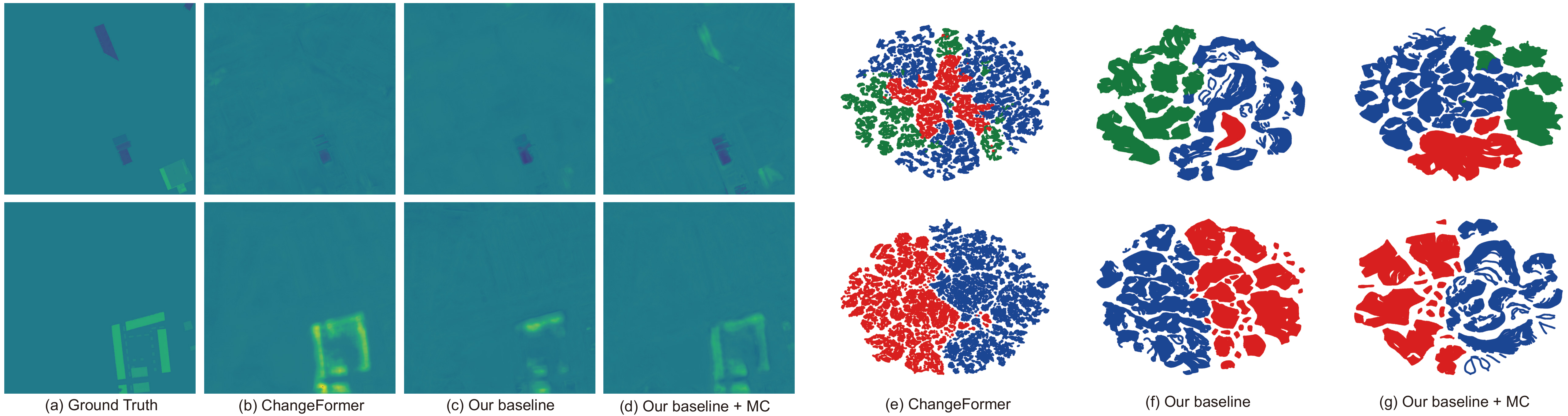}
	\caption{The feature space of last decoder layer visualized by t-SNE. The red indicates a new-built building pixel, green indicates a demolished building pixel, and blue indicates an unchanged pixel.}
	\label{fig:vis_tsne3}
\end{figure*}

{Figure \ref{fig:large_3d} depicts the large-scale change detection results. Figure \ref{fig:vis_tsne1} and \ref{fig:vis_tsne3} visualize the multimodal feature from the encoder and decoder layers. The baseline of our model is the one that does not have the multitask consistency constraint (MC). Given that our Transformer backbone is a streamlined variant of ChangeFormer, we include ChangeFormer's results for comparative analysis. Our baseline differs from ChangeFormer in having fewer parameters and incorporating a cross-modal fusion module. In Figure \ref{fig:vis_tsne1}, the multimodal features from DSM (blue points) and image (red points) are roughly separated into two parts with some overlapping. From left to right, the red cluster is more and more compact. In Figure \ref{fig:vis_tsne3}, the three distinct classes are represented with specific colors in the output: demolished buildings are marked in red, newly-built buildings in blue, and the background in green. 
	Our method, when compared to both ChangeFormer and our baseline model that did not incorporate consistency constraints, demonstrates an improvement in the feature space representation. Specifically, it shows better intra-class consistency and inter-class separation.}

\subsection{Ablation study}

\begin{table*}[!b]
	\caption{Ablation study about the impact of multitask consistency (+MC) on height change predicting branch (+3d) and semantic change detection (+2d) in single-task or multitask scenarios. {The '-' symbol indicates that the ablated model does not contain the corresponding component to get the metric outputs.}}
	\label{table-ablation_semantic}
	\vspace{2mm}
	\centering
	\scalebox{1.0}{
		\begin{tabular}{ccc|cccc|ccccc}
			\toprule
			\multicolumn{3}{c|}{\text {Settings}} & \multicolumn{4}{c|}{\text {Semantic metrics}} & \multicolumn{5}{c}{\text {Height metrics}} \\
			\hline 
			\text{+2d} & +3d  & {+MC} & ${IoU_D}$ & ${IoU_N}$ & mIoU & \text{F1score}  & {RMSE} & {MAE} & CRMSE & cRel & ZNCC  \\
			\hline 
			\checkmark & & &  43.76&39.10&41.43&58.55&-&-&-&-&- \\
			& \checkmark& &-&-&-&-&1.460&0.301&8.289&2.075&0.311\\
			\checkmark & \checkmark  & & 39.05&37.31&38.18&55.26&1.367&0.358&8.875&1.922&0.311 \\
			& \checkmark & \checkmark  & -&-&-&-&1.273&0.397&8.317&2.711&0.379 \\
			\checkmark&\checkmark&\checkmark&44.29&40.90&42.59&59.72&1.267&0.290&8.281&1.900&0.394\\
			
			\bottomrule
	\end{tabular}}
\end{table*}

This section explores the {impact} of the proposed multitask consistency for semantic and height change detection. Initially, {w}e demonstrate that the implicit information, shared at the backbone stage, yields benefits for height change detection but {introduces} challenges for semantic change detection.

\begin{figure*}[!t]%
	\centering
	\includegraphics[width=1.0\textwidth]{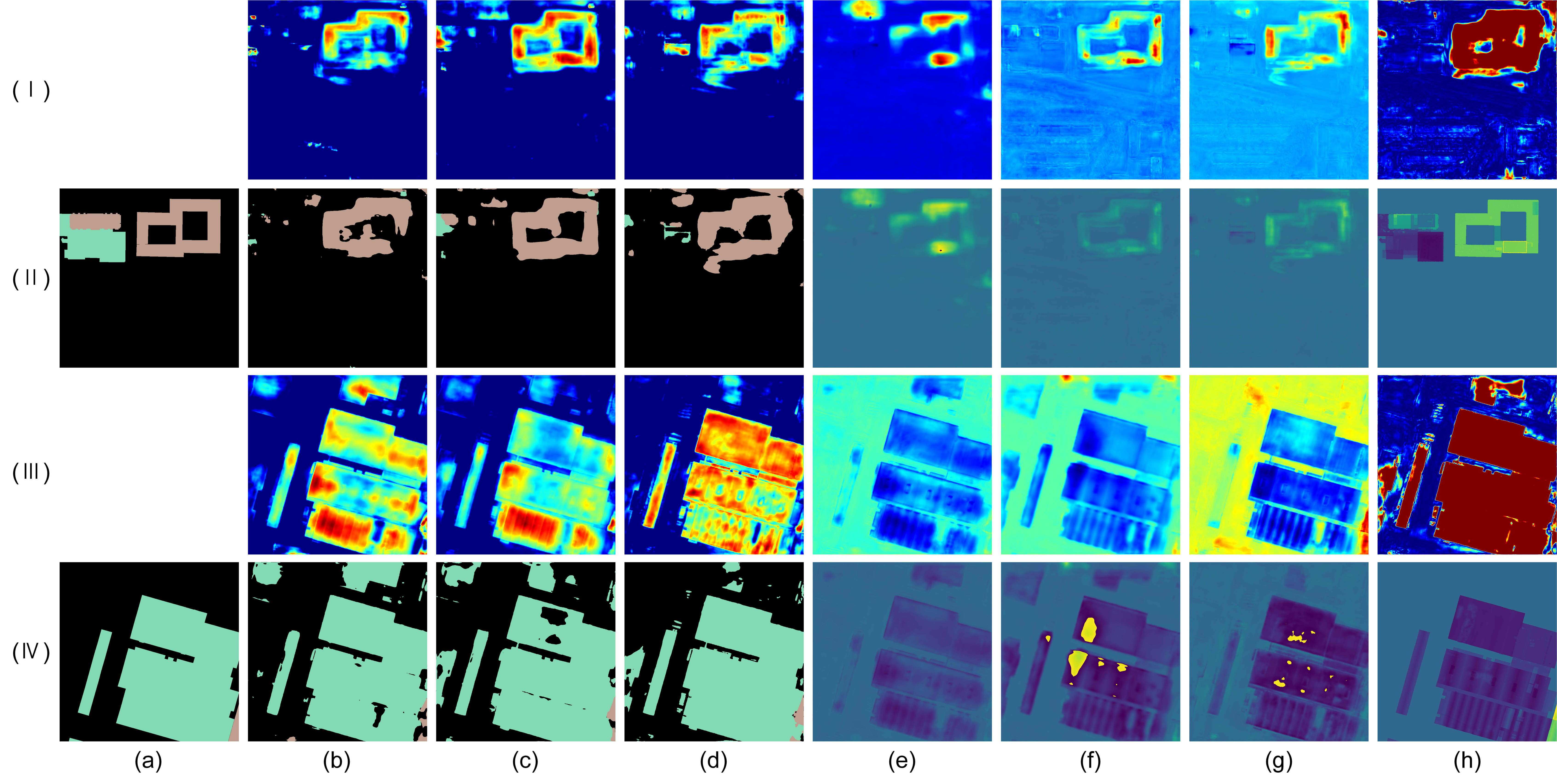}
	\caption{Visual examples highlighting the improvement via multitask interaction. (a) semantic ground truth; (b)(e) single-task results and corresponding attention maps; (c)(f) multitask results and corresponding attention maps; (d)(g) multitask results and corresponding attention maps with consistency constraint; (h) attention maps for soft thresholding layer (odd-numbered rows) and ground truth height changes (even-numbered rows).}\label{fig_ablation1}
\end{figure*}

From the results of row 1 and row 3 in Table \ref{table-ablation_semantic}, an evident decline in semantic change detection can be observed. Conversely, the height metric results from rows 2 and 3 demonstrate the utility of semantic hints for estimating in enhancing the height changes {estimation}, even with only implicit shared information in the common backbone. Figure \ref{fig_ablation1}(b)(c)(e)(f) illustrates some visual examples wherein the semantic branch exhibits improved recovery of height changes. However, it tends to introduce {additional} noise in the background regions. From the odd-numbered rows of columns (c) and (f), we can observe that the attention regions corresponding to the semantic output closely resemble height changes. This suggests a strong coupling between their learned representations, which is the reason why the inclusion of the height branch poses challenges for semantic change detection, resulting in suboptimal performance in both tasks, as demonstrated in Figure \ref{fig_ablation2}(b)(c)(e)(f). To alleviate the problem, we designed the pseudo-change branch for two purposes:

\begin{figure*}[!t]%
	\centering
	\includegraphics[width=1.0\textwidth]{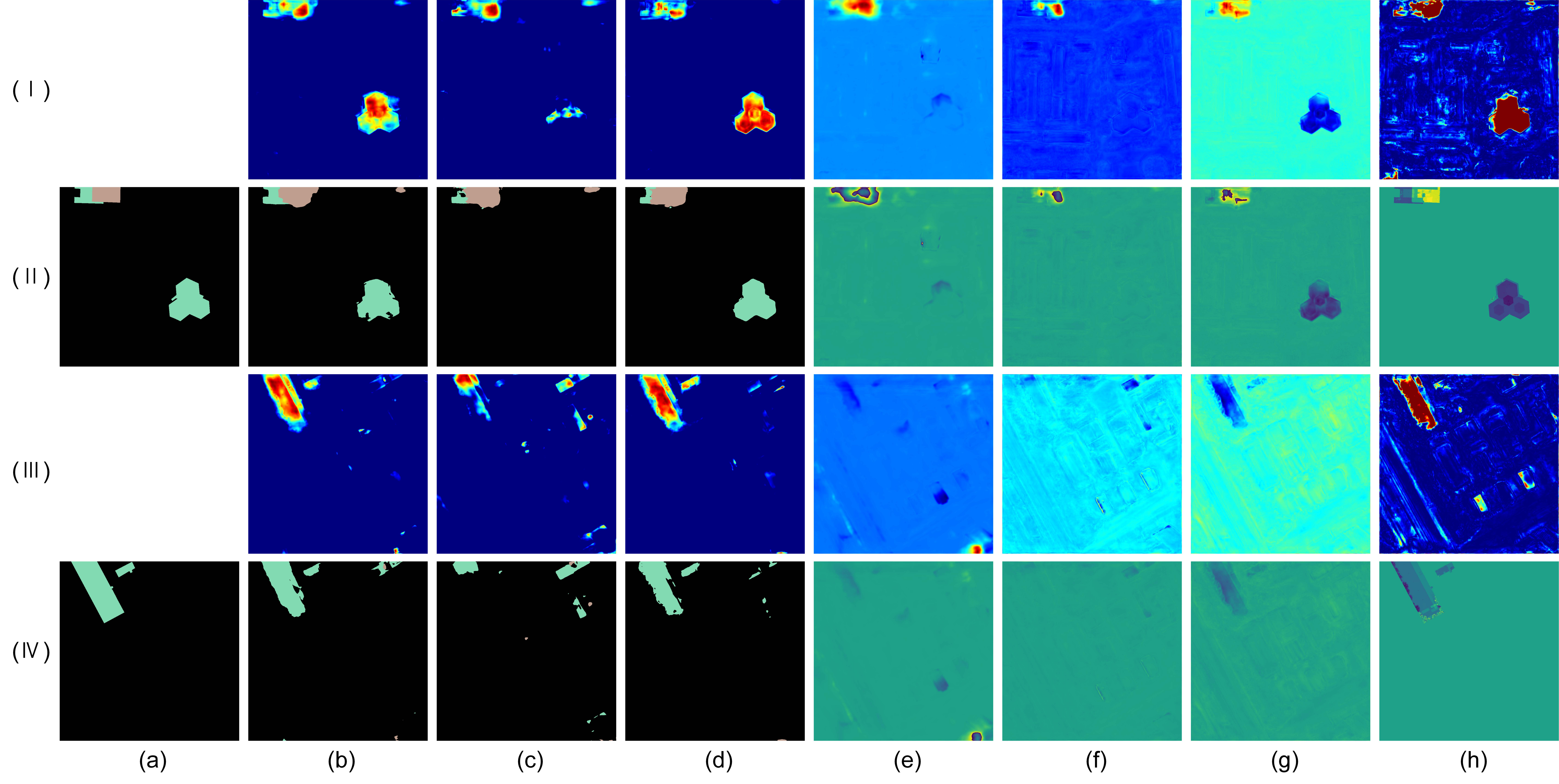}
	\caption{Visual examples highlighting the decline due to multitask branches. (a) semantic ground truth; (b)(e) single-task results and corresponding attention maps; (c)(f) multitask results and corresponding attention maps; (d)(g) multitask results and corresponding attention maps with consistency constraint; (h) attention maps for soft thresholding layer (odd-numbered rows) and ground truth height changes (even-numbered rows).}\label{fig_ablation2}
\end{figure*}

1) {In the absence of semantic change hints}, it serves as a sub-complete semantic map to assist in the height change detection, resulting in an augmented height detection branch. Comparing row 2 to row 4 in Table \ref{table-ablation_semantic}, we found that the semantic hint from the pseudo change branch is even better than the original semantic branch. We speculate that it was attributed to the explicit soft thresholding process, which brings stronger prior about multitask relationship than the limited hints from shared backbone. The odd-numbered rows of the final column in Figure \ref{fig_ablation1} and \ref{fig_ablation2} provide a visualization of learned representation in the soft thresholding layer, which accurately locates some edge cases that were missed by the original multitask scenario.

\begin{figure*}[!b]%
	\centering
	\includegraphics[width=1.0\textwidth]{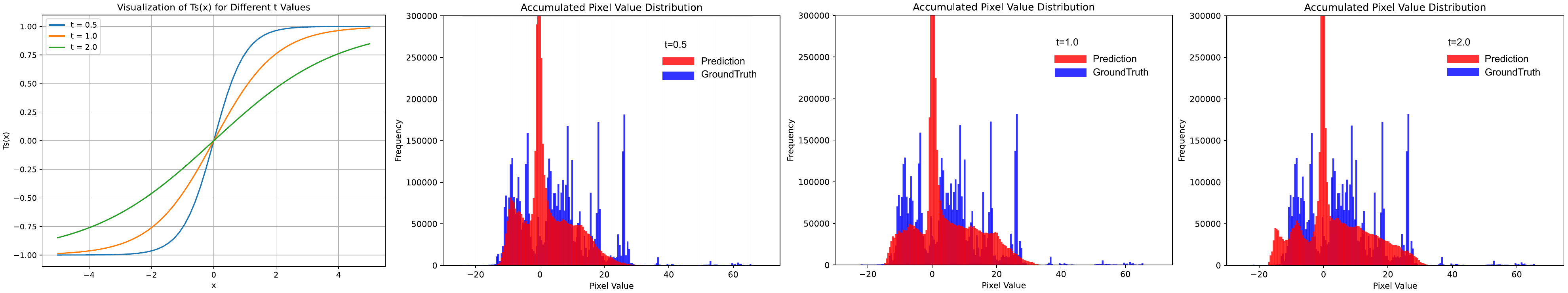}
	\caption{The impact of different $t$ values on height change detection.}\label{fig_heightbars_difft}
\end{figure*}

2) {With both semantic and height change branches in place}, the pseudo change branch {fosters} a consistent relationship between the two tasks. 
Interestingly, the final row of Table \ref{table-ablation_semantic} demonstrates that our consistency strategy has not only mitigated the multitask conflicts but also encouraged further improvements via explicit multitask interaction, as shown in Figure \ref{fig_ablation1}(d)(g) and \ref{fig_ablation2}(d)(g). Furthermore, our consistency strategy can be seamlessly applied to the other change detection methods and yields promising improvements, as shown in Table \ref{table-mc_for_sotas}.

\begin{table*}[!t]
	\caption{The impact of our consistency strategy on the other methods, including one CNN-based and two Transformer-based change detection models. The {\bf bold} font indicates better results.}
	\label{table-mc_for_sotas}
	\vspace{2mm}
	\centering
	\scalebox{1.0}{
		\begin{tabular}{c|ccccc|cc}
			\toprule 
			\multirow{2}{*}{\text{Settings}}&\multicolumn{5}{c|}{\text{Height metrics}}&\multicolumn{2}{c}{\text{Semantic metrics}}\\
			
			\cline{2-8}
			
			~&\text{RMSE}&\text{MAE}&\text{cRMSE}&\text{cRel}&\text{ZNCC}& mIoU$\uparrow$& F1-score$\uparrow$\\
			
			\hline
			
			\text{FC-siamese}&1.461&{\bf 0.309}&8.995&{\bf 1.506}&0.373&27.97&43.72\\
			
			+\text{multitask consistency}&{\bf 1.398}&0.353&{\bf 8.655}&1.873&{\bf 0.395}&{\bf 28.85}&{\bf 44.78}\\
			
			\hline
			
			\text{ChangeFormer}&{1.343}&{0.402}&{8.204}&{2.485}&{ 0.394}&{40.17}&{57.31}\\
			
			+\text{multitask consistency}&{\bf 1.317}&{\bf 0.297}&{\bf 8.085}&{\bf 1.825}&{\bf 0.447}&{\bf 41.37}&{\bf 58.39}\\
			
			\hline
			
			\text{MTBIT}&1.457&0.400&{\bf 8.563}&1.987&{\bf 0.345}&30.72&{46.88}\\
			
			+\text{multitask consistency}&{\bf 1.373}&{\bf 0.343}&8.788&{\bf 1.639}&0.281&{\bf 32.44}&{\bf 48.90}\\
			\bottomrule
	\end{tabular}}
\end{table*}

To further investigate the impact of our consistency constraint on height change detection, We {adjusted} the temperature parameter of equation \ref{eq_softthresh}. Figure \ref{fig_heightbars_difft} shows that a smaller temperature value leads to a sharper transition near zero and greater overlap between the ground truth and predicted height change. This is because the sharp transition near zero suppresses background noise where height is unchanged, allowing more attention {to} changed regions. However, this parameter acts as a double-edged sword; a too-small value means that background noise is more likely to be mistaken for a change target. Therefore, in our experiments, we set $t$ to 0.5 as the final setting.

\begin{table}[!thp]
	
	\caption{The influence of different combinations of hyperparameters.}
	\label{table-hypercomb}
	\vspace{2mm}
	\centering
	\begingroup
	
	\scalebox{1.0}{
		\begin{tabular}{c|cc|cc}
			\toprule
			\multirow{2}*{Hyperparameters} & \multicolumn{2}{c|}{semantic metrics} & \multicolumn{2}{c}{height metrics}  \\
			\cline{2-3} 
			\cline{4-5}
			~ & mIoU & F1-score &RMSE  & MAE\\
			\hline
			
			$\lambda_1=0.2,\lambda_2=0.6,\lambda_3=0.6$&41.42&58.53&1.263&0.289\\
			$\lambda_1=0.6,\lambda_2=0.2,\lambda_3=0.6$&43.24&60.22&1.302&0.388\\
			$\lambda_1=0.6,\lambda_2=0.6,\lambda_3=0.6$&42.38&59.49&1.289&0.301\\
			$\lambda_1=0.2,\lambda_2=0.2,\lambda_3=0.6$& 42.59& 59.72& 1.267&0.290 \\
			\bottomrule
	\end{tabular}}
	\endgroup
\end{table}

{By setting the $\lambda_3$ parameter of the semantic branch to 0.6, we explore various parameter combinations within Table \ref{table-hypercomb}.}
{In the first row, equating the loss weights for the height and semantic change branches results in a degradation of semantic change detection performance, with minimal improvement observed in the height change metrics. Conversely, increasing $\lambda_1$ for the pseudo change branch enhances the semantic change detection metrics, underscoring the task's emphasis, albeit at the cost of diminished height change metrics. When $\lambda_1$ to $\lambda_3$ are equalized, there's a notable improvement in height metrics compared to the configuration in row 2. Nonetheless, this increased weight on the pseudo change branch continues to adversely impact the height change detection performance relative to the baseline established in row 1. For further details, please consult Table 5 in our revised manuscript.}

\subsection{Discussion}

The experiment results strongly emphasize the multitask conflicts between 2D semantic change detection and 3D height estimation. Specifically, the semantic change exhibits a distinct boundary that aids in pinpointing changes compared to height change estimation. A similar phenomenon has been documented in \cite{zhu2020edge}, which highlights that object boundaries are more readily discerned from segmentation labels than from depth maps. From Figure \ref{fig_ablation1} and \ref{fig_ablation2}, we could observe the inherent consistency between height change and the activation map of semantic change within a single testing example. This suggests that the decreased performance of the multitask setting is mainly caused by the height branch. As denoted in \cite{liu2021conflict}, jointly addressing semantic segmentation and depth estimation tasks tends to yield suboptimal results compared to single-task settings.

Our proposed multitask consistency constraint {links} the height change branch and pseudo-change branch via soft-thresholding. By minimizing the disparity between pseudo-change and semantic change, we enable gradient interaction from the semantic change map to the height change map. This establishes a coherent objective for the multitask branches and ultimately enhances the performance of both tasks, as evidenced in our experiments.

{Our approach incorporates DSM data derived from point cloud data. Looking ahead, we aim to enhance change detection between point cloud and image data. This will involve refining representation learning methods that learn visual representations from multimodal data (point clouds and images) without extensive labeled datasets \cite{wu2023self, wu2023sacf}. Additionally, the techniques of multi-scale feature fusion \cite{wu2023panet} between point cloud and image data is pivotal for effective multimodal change detection.}

\section{Conclusion}\label{sec5}

The prevailing direction in change detection research is toward achieving higher frequency, finer granularity, and increased dimensionality. However, there exists a noticeable gap in the literature {about} multimodal and cross-dimensional change detection. In this paper, we presented a novel pipeline for detecting height and semantic change simultaneously from DSM-to-image multimodal data. We revealed that the leading change detection methods, including CNN-based and Transformer-based methods, struggled with the conflicts of multitask change detection. We proposed a Transformer-based network equipped with multitask consistency constraint, which achieves the best semantic and height change detection performance with limited model complexity. We found that the consistency strategy with a small temperature parameter is able to suppress the background noise and leads to sharper results in change regions. It can be also seamlessly employed to the other change detection methods and produce promising improvements. {Our dataset and model are poised to become foundational benchmarks for future research within the remote sensing change detection community. This work paves the way for a range of intriguing research avenues, such as the exploration of more fine-grained semantic categorizations based on varying scales of height changes, and the integration of currently dominant large pre-trained models into our framework.}

{However, several unresolved questions persist, including both data and algorithmic aspects. Firstly, we have initiated a dataset expansion plan, necessitating enhanced automation in our annotation workflow to accommodate large-scale applications. Furthermore, as aforementioned in Section \ref{sec_compare}, we need to delve deeper into addressing the substantial class imbalance issue with various strategies, including refining training metrics, implementing data augmentation techniques, and optimizing the architecture. Additionally, leveraging the power of state-of-the-art large pre-trained models could lead to a more efficient training process and improved generalization.}


%

\newpage

\vfill

\end{document}